\documentclass{article}
\usepackage{ijcai25}

\usepackage{times}
\usepackage{color}
\usepackage{soul}
\usepackage{url}
\usepackage[hidelinks]{hyperref}
\usepackage[utf8]{inputenc}
\usepackage[small]{caption}
\usepackage{graphicx}
\usepackage{amsmath}
\usepackage{amsthm}
\usepackage{booktabs}
\usepackage{algorithm}
\usepackage{algorithmic}
\usepackage[switch]{lineno}
\usepackage{makecell}
\usepackage{multirow}
\usepackage{threeparttable}
\usepackage{amssymb}
\usepackage{enumitem}

\title{Neuromorphic Sequential Arena: A Benchmark for Neuromorphic Temporal Processing}

\author{
Xinyi Chen$^{1,}$\footnotemark[2]
\and
Chenxiang Ma$^{1,}$\footnotemark[2]\and 
Yujie Wu$^{2}$\and
Kay Chen Tan$^{1,3}$\And
Jibin Wu$^{1,2,3}$ 
\footnotemark[1]\\
\affiliations
$^1$Department of Data Science and Artificial Intelligence, The Hong Kong Polytechnic University\\
$^2$Department of Computing, The Hong Kong Polytechnic University\\
$^3$Research Center of Data Science and
Artificial Intelligence, The Hong Kong Polytechnic University\\
\emails
\{xinyi-97.chen,~chenxiang.ma\}@connect.polyu.hk,
wu-yj16@tsinghua.org.cn,
\{kctan,~jibin.wu\}@polyu.edu.hk
}
\renewcommand{\thefootnote}{\fnsymbol{footnote}}
\begin{document}

\maketitle
\begingroup
\renewcommand\thefootnote{}\footnotetext{This paper has been accepted to IJCAI 2025.}
\addtocounter{footnote}{0}
\endgroup

\footnotetext[1]{Corresponding author: Jibin Wu.} \footnotetext[2]{These authors contributed equally to this work.}

\begin{abstract}
Temporal processing is vital for extracting meaningful information from time-varying signals. Recent advancements in Spiking Neural Networks (SNNs) have shown immense promise in efficiently processing these signals. However, progress in this field has been impeded by the lack of effective and standardized benchmarks, which complicates the consistent measurement of technological advancements and limits the practical applicability of SNNs. To bridge this gap, we introduce the Neuromorphic Sequential Arena (NSA), a comprehensive benchmark that offers an effective, versatile, and application-oriented evaluation framework for neuromorphic temporal processing. The NSA includes seven real-world temporal processing tasks from a diverse range of application scenarios, each capturing rich temporal dynamics across multiple timescales. Utilizing NSA, we conduct extensive comparisons of recently introduced spiking neuron models and neural architectures, presenting comprehensive baselines in terms of task performance, training speed, memory usage, and energy efficiency. 
Our findings emphasize an urgent need for efficient SNN designs that can consistently deliver high performance across tasks with varying temporal complexities while maintaining low computational costs.  
NSA enables systematic tracking of advancements in neuromorphic algorithm research and paves the way for developing effective and efficient neuromorphic temporal processing systems. 
\end{abstract}

\section{Introduction}

Temporal processing is fundamental for intelligent systems to interpret time-varying sensory signals, facilitating accurate and timely decision-making in dynamic environments. Neuromorphic computing holds immense potential for processing these signals with ultra-low energy consumption and low latency. Spiking Neural Networks (SNNs), inspired by the computational principles of biological brains \cite{maass1997networks}, serve as a cornerstone of neuromorphic computing. While SNNs have achieved performance on par with traditional Artificial Neural Networks (ANNs) in many static image classification tasks, demonstrating significantly improved energy efficiency and reduced latency~\cite{davies2018loihi,pei2019towards,ma2024darwin3,yang2024vision}, their capacities in processing temporal signals remain inferior to those of ANNs. Recently, numerous SNN approaches have been proposed, making substantial progress in bridging this gap. These advancements include enriching neural dynamic heterogeneity \cite{yin2021accurate,zheng2024temporal}, increasing neuronal structural complexity \cite{zhang2024tc,he2024network,sun2024delayed}, and developing temporal parallelization methods for improved training efficiency \cite{fang2024parallel,chen2024pmsn}.

Despite this progress, advancements in this research area are hindered by the absence of effective benchmarks. Existing {SNN}
benchmarks predominantly focus on visual classification \cite{lecun1998gradient,li2017cifar10,amir2017low} and keyword spotting \cite{warden2018speech,cramer2020heidelberg} tasks. While these benchmarks have been valuable over the past decade in advancing neuromorphic computing research, they play a limited role in fostering developments in neuromorphic temporal processing due to three primary limitations. First, the model performance on the current SNN benchmarks is largely saturated, and these benchmarks fail to capture the rich temporal dynamics inherent in real-world signals, which typically encompass multiple timescales. Second, existing benchmarks do not adequately represent the diverse range of temporal processing scenarios that closely align with the interests and objectives of neuromorphic computing research. Third, inconsistent comparisons across studies and the absence of evaluations on critical efficiency metrics result in biased assessments of model practicality.
%Third, there lacks a standardized criterion for assessing the effectiveness of benchmarks in evaluating SNNs’ temporal processing capacities.
 
To address these limitations, we propose a comprehensive benchmark called \textbf{Neuromorphic Sequential Arena (NSA)}, designed to establish an effective, versatile, and application-oriented evaluation framework for neuromorphic temporal processing. 
Firstly, NSA encompasses a broad range of tasks that reflect real-world temporal processing scenarios across application areas of significant interest to the neuromorphic research community, including human-computer interaction, speech processing, robotics, and biomedical applications. Furthermore, to ensure that these tasks possess an adequate level of temporal complexity suitable for SNNs, we introduce a novel tool for analyzing the temporal dependencies required to address a given task, referred to as the Segregated Temporal Probe (STP). This tool has been employed to validate the effectiveness of both commonly used neuromorphic datasets and the proposed NSA in benchmarking the temporal processing capacity of SNNs. Finally, using NSA, we conduct a comprehensive comparative study of recently introduced spiking neuron models and neural architectures in terms of task performance, training speed, memory usage, and energy efficiency. By providing a side-by-side comparison of these baselines, NSA serves as a valuable foundation for understanding the current status and practicality of existing methods. Collectively, NSA is designed as an evolving framework capable of integrating emerging neuromorphic algorithms, thereby fostering advancements toward more effective and efficient neuromorphic temporal processing systems.

\section{Neuromorphic Sequential Arena (NSA)}
The NSA comprises seven tasks that span a diverse array of real-world scenarios relevant to neuromorphic research. These tasks, characterized by varying levels of temporal complexity, serve as an effective benchmark for assessing the temporal processing capacities of different SNN approaches. In the following, we detail our design principles, task formulation, and STP tool used to assess task effectiveness.
\subsection{Design Principles}
This section outlines the five foundational principles for the NSA design:
\begin{itemize}[left=0pt]
    \item \textbf{Neuromorphic {relevance}.} Tasks should highlight the advantages of neuromorphic solutions, such as energy efficiency, low latency, and robustness.
    \item \textbf{Temporal complexity.} Tasks should reflect rich temporal dynamics inherent in real-world scenarios, requiring models to establish temporal dependencies across multiple timescales.
    \item \textbf{Challenging.} Tasks should exhibit an adequate level of difficulty, highlighting distinguishable performance among existing SNN approaches while offering significant opportunities for improvement. 
    \item \textbf{Training resource.} Tasks should account for training time and GPU resource constraints to ensure accessibility for both researchers and practitioners. In particular, for SNN approaches that rely on temporally serial simulation, it is crucial that the training can be completed within a reasonable amount of training time.
    \item \textbf{Application versatility.} Tasks should encompass a wide range of real-world application scenarios with distinct requirements and data characteristics, bridging the gap between theoretical advancements and practical applications.
\end{itemize}
\subsection{Tasks}
\begin{table*}[h!]
\centering
\caption{Summary of tasks characteristics, evaluation metrics, and other configurations in the proposed NSA benchmark.}
\resizebox{0.80\textwidth}{!}{
\begin{tabular}{@{}llcccrr@{}}
\toprule
\textbf{Task} & \textbf{Dataset}  & \textbf{Sequence length} & \textbf{Metric} & \textbf{Backbone model}  & \textbf{Training samples} & \textbf{Testing samples} \\ \midrule
AL & AL  & User-defined & Acc. & MLP & 50,000 & 5,000 \\ 

HAR & WISDM   & 200 & Acc. & MLP  & 26,048 & 6,512 \\ 

EEG-MI & OpenBMI  & 500 & Acc. & MLP  & 17,280 & 4,320 \\ 
SSL & SLoClas & 500 & Acc. & MLP  & 37,426 & 8,969 \\ 
ALR & DVS-Lip  & 200 & Acc. & MLP & 14,896 & 4,975 \\ 
AD & N-DNS  & 751/3,751 & SI-SNR & Spiking-FullSubNet  & 60,000 & 341 \\ 
ASR & AISHELL & 76–505 & CER & VGG-MLP  & 360,294 & 7,176 \\ \bottomrule
\end{tabular}}
\label{tab:task_sum}
\end{table*}

In the following, we will introduce the suite of tasks included in the proposed NSA benchmark, meticulously designed following the principles outlined above. We provide a comprehensive overview of these tasks, including detailed task descriptions, application scenarios, and specific capacities they evaluate. The task characteristics, evaluation metrics, and dataset configurations are summarized in Table~\ref{tab:task_sum}. Detailed preprocessing techniques for these tasks are provided in the Supplementary Materials. To facilitate benchmarking efforts, we provide a comprehensive \textbf{open-source library}\footnote{The source code and supplementary materials are publicly available at \url{https://github.com/liyc5929/neuroseqbench}.} that allows seamless integration of novel spiking neuron models and neural architectures, and ensures consistent evaluations across different methods.

\subsubsection{Autonomous localization (AL)}
Due to the inherent characteristics of low latency and high energy efficiency, neuromorphic systems present considerable potential for robotic control. 
A fundamental challenge in this domain is predicting the current state of the robot after executing a sequence of actions, which is crucial for overcoming sensory feedback delays and promptly adapting to external perturbations. 
Addressing this challenge requires a neuromorphic system to establish long-term temporal dependencies, effectively modeling the intricate relationship between past actions and the present state.

Given the above requirement and the significance of this task, we propose a new synthetic dataset called AL. AL simulates a scenario in which a mobile robot must determine whether it is positioned on the left or right plane following a sequence of ordered actions, including `turn left', `turn right', `go straight', and `stop'. This constitutes a binary classification task with customized sequence lengths and action distributions, facilitating an effective evaluation of the capacity of SNNs to establish temporal dependencies across various timescales.

% \begin{figure}[h]
% \centering
% \includegraphics[width=0.9\linewidth]{fig/ALL.pdf}
% \caption{Demonstration of the autonomous localization task.} 
% \label{fig: all}
% \end{figure}

\subsubsection{Human activities recognition (HAR)}
The HAR task focuses on identifying human activity patterns from time-series data collected by wearable sensors. This task represents a significant application of neuromorphic computing, encompassing real-time robotic locomotion control, pose estimation, and surveillance, where accurate and efficient temporal processing is essential. HAR requires the modeling of fine-grained temporal trajectories of human activities embedded within noisy sensor signals. Therefore, HAR is particularly well-suited for evaluating the capacity of SNNs to capture complex short-term temporal dependencies and to demonstrate their robustness against unpredictable noise conditions.

In this task, we utilize data samples from the WISDM dataset~\cite{weiss2019wisdm}, which includes gyroscope data collected from a smartwatch. The data is recorded at 20 Hz and segmented into 10-second intervals, resulting in a sequence length of 200. This task is intentionally challenging, requiring models to accurately classify 18 actions that may exhibit subtle differences, such as `eating chips' or `eating pizza'.

\subsubsection{Electroencephalogram motor imagery (EEG-MI)}
We further evaluate the effectiveness of SNNs in brain-computer interfaces and real-time cognitive monitoring through the EEG-MI task. This task focuses on decoding motion imagery from EEG sequences, which is particularly challenging due to the sparse, noisy, and highly dynamic nature of EEG signals. Therefore, it serves as an effective benchmark for assessing the capacity of SNNs to capture both temporal and spatial dependencies with a high degree of robustness.

Specifically, we utilize motion imagery data from the OpenBMI dataset \cite{lee2019}, which comprises 62-channel EEG recordings collected from 52 subjects at a sampling rate of 1 kHz. The model is tasked with performing a binary classification aimed at distinguishing between imagery trials of left- and right-hand grasping. To ensure manageable training costs, we downsample the sequence length to 500. To mitigate the risk of overfitting due to subject-specific bias, we employ a cross-trail validation approach, in which all samples are randomly partitioned into training and testing sets. 

\subsubsection{Sound source localization (SSL)}
Identifying the origin of a sound source within a noisy environment is a crucial survival skill for animals and holds significant importance for applications such as hearing aids and robotics. In this task, we aim to evaluate the temporal processing capacity and noise robustness of SNN approaches. To this end, we adopt the SLoClas dataset \cite{qian2021sloclas}, which contains 4-channel audio recordings from a single sound source positioned at azimuth angles ranging from 0$^{\circ}$ to 360$^{\circ}$ in 5$^{\circ}$ increments, resulting in a 72-class classification problem. Additionally, directional background noise fragments are introduced to the raw audio signals at a challenging signal-to-noise (SNR) ratio of 0. The resulting audio signals are segmented into sequences of length 500, then fed directly into the SNNs without any spectral preprocessing. This task poses a specific challenge for temporal processing, as the model must learn to map the temporal delays between different audio channels to their corresponding azimuth angles in the presence of background noise.

\subsubsection{Automatic lip-reading (ALR)}
ALR aims to recognize spoken words from a speaker's lip movements and plays a pivotal role in various real-world applications, such as video surveillance and speech recognition in noisy environments. Dynamic Vision Sensor (DVS) cameras, characterized by high dynamic range and low latency, have emerged as ideal visual front-ends for capturing the fine-grained lip movements essential for the ALR task. Their event-based outputs are particularly well-suited for neuromorphic systems.

In this task, we adopt the DVS-Lip dataset~\cite{tan2022multi}, which consists of 100 spoken word classes captured by the DAVIS346 event camera from 40 individuals. The task is particularly challenging, as the training and testing sets include different individuals, requiring the SNN model to generalize across unseen speakers. Moreover, the output is decoded solely from the spiking activities generated at the final time step, requiring models to extract and retain crucial spatiotemporal features over an extended period. To ensure manageable training costs, each sample is segmented into 200 temporal bins and center-cropped to a resolution of $88\times88$ pixels.

\subsubsection{Audio denoising (AD)}
Removing noise from received audio signals to enhance overall quality is critical for many edge applications, such as hands-free communication and hearing aids. Given the low-power and real-time processing requirements of these tasks, neuromorphic solutions present a highly promising approach. Notably, among all tasks within the proposed NSA benchmark, AD stands out as a regression problem, providing a unique opportunity to evaluate the expressive power of SNN models in capturing subtle and continuous temporal variations.

In this task, we utilize the publicly available Intel Neuromorphic Deep Noise Suppression (N-DNS) Challenge dataset~\cite{timcheck2023intel}. This dataset comprises clean speech samples in multiple languages (i.e., English, German, French, Spanish, and Russian) as well as noise samples collected from diverse acoustic environments. We employ the official synthesizer script to generate a 495-hour training subset and a 5-hour validation subset. All audio samples are synthesized at a 16 kHz sampling rate and segmented into a uniform length of 30 seconds. We perform audio preprocessing steps in accordance with the winning entry of the N-DNS Challenge~\cite{hao2024towards} and adopt their model architecture as the default. For model evaluation, we adopt the Scale-Invariant Signal-to-Noise Ratio (SI-SNR) metric.

\subsubsection{Automatic speech recognition (ASR)}
ASR transcribes spoken language from audio into text, serving as a foundational technology for various applications, including voice assistants, transcription services, and speech translation tools. This task presents significant challenges for SNNs due to varying sequence lengths, speaker characteristics, and acoustic conditions. 

In this task, we utilize the AISHELL dataset~\cite{bu2017aishell}, an open-sourced Mandarin speech corpus comprising approximately 170 hours of speech data from 400 speakers, encompassing a wide range of accents and speaking styles. Its moderate data scale and high speaker diversity render it an effective testbed for assessing the model's capacity to handle linguistic and acoustic variations. Model performance is evaluated using the Character Error Rate (CER).

\subsection{Segregated Temporal Probe (STP)}

\begin{figure}[t]
\centering\includegraphics[width=0.98\linewidth, trim = 0 0 0 0, clip]{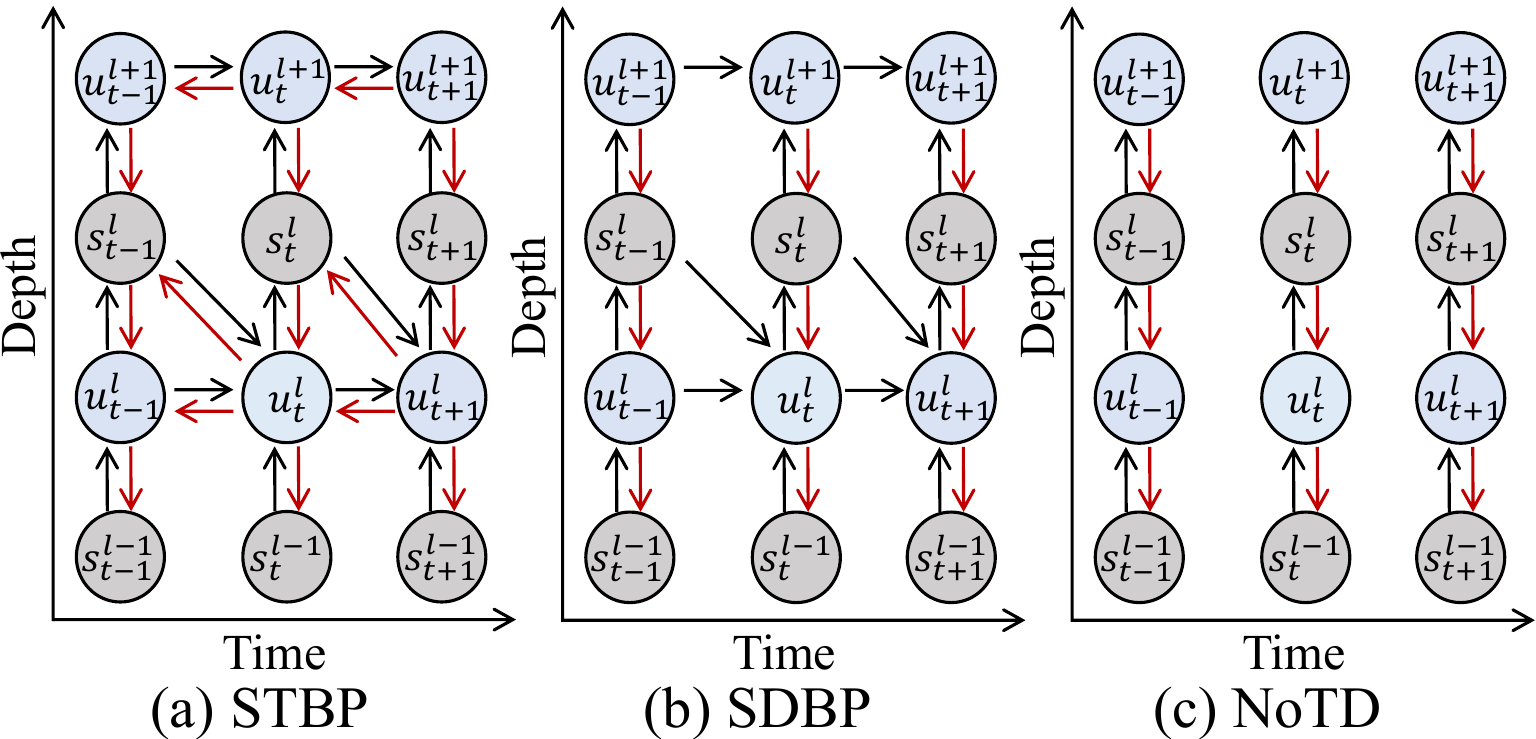}
\caption{Comparison of the three training algorithms in STP.} 
\label{fig: stp}
\end{figure}
To elucidate the limited effectiveness of commonly used neuromorphic benchmarks in evaluating the temporal processing capacity of SNNs and to demonstrate the efficacy of NSA, we introduce STP. STP assesses the contributions of establishing temporal dependencies by quantifying the impact of isolating forward and backward temporal processing pathways of spiking neurons on task performance. As illustrated in Figure~\ref{fig: stp}, STP consists of three evaluation modules, each corresponding to a specific training algorithm: Spatio-Temporal Backpropagation (STBP)~\cite{wu2018spatio}, Spatial Domain Backpropagation (SDBP), and No Temporal Domain (NoTD). 
Details of these three algorithms are outlined below, using the Leaky Integrate-and-Fire (LIF) neuron model~\cite{burkitt2006review} as an example.

The LIF neuron accumulates input spikes $\boldsymbol{s}^{l-1}[t]$ from the preceding layer $l-1$ into its membrane potential $\boldsymbol{u}^{l}[t]$. When $\boldsymbol{u}^{l}[t]$ surpasses a threshold $V_{\mathrm{th}}$, it triggers a spike and subsequently resets to the resting potential. The dynamics of a LIF neuron can be formulated as:
\begin{align}
\boldsymbol{u}^{l}[t] &= \underbrace{\lambda  \boldsymbol{u}^{l}[t - 1] (1 - \boldsymbol{s}^{l}[t-1])}_{\text{Forward temporal propagation}} +\, \boldsymbol{W}^{l} \boldsymbol{s}^{l-1}[t], \label{eq:mem_update}\\
\boldsymbol{s}^{l}[t]  &= \Theta (\boldsymbol{u}^{l}[t] - V_{\mathrm{th}}),
\end{align}
where $\lambda$ determines the decay rate of $\boldsymbol{u}^{l}[t]$ over time, $\boldsymbol{W}^{l}$ is the synaptic weight matrix, and $\Theta(\cdot)$ is the Step function. The recursive update in Eq.~\eqref{eq:mem_update} allows the LIF neuron to iteratively propagate temporal information from $\boldsymbol{u}^{l}[t-1]$ to $\boldsymbol{u}^{l}[t]$, enabling the integration of temporal information over time.

\textbf{STBP} preserves this temporal propagation in both forward and backward passes, where the gradient of the loss with respect to weights is computed as:
\begin{align}
\!\frac{\partial \mathcal{L}}{\partial \boldsymbol{W}^{l}} &= \sum_{t=1}^{T}\frac{\partial \mathcal{L}}{\partial \boldsymbol{u}^{l}[t]} \frac{\partial \boldsymbol{u}^{l}[t]}{\partial \boldsymbol{W}^{l}} = \sum_{t=1}^{T}{\boldsymbol{\delta}^{l}[t]}^{\!\top} {\boldsymbol{s}^{l-1}[t]}^{\!\top}, \label{eq:weight}
\\[1.5ex]
\boldsymbol{\delta}^{l}[t]\!&=\underbrace{\!
\boldsymbol{\delta}^{l}\![t\!+\!1]\frac{\partial \boldsymbol{u}^{l}\![t+1]}{\partial \boldsymbol{u}^{l}\![t]}}_{\text{Backward temporal propagation}} + \boldsymbol{\delta}^{l+1}\![t] \frac{\partial \boldsymbol{u}^{l+1}\![t]}{\partial \boldsymbol{u}^{l}\![t]},\!\! 
\label{eq:grad_m}
\end{align}
where $\boldsymbol{\delta}^{l}[t]\triangleq\frac{\partial \mathcal{L}}{\partial \boldsymbol{u}^{l}[t]}$, which is the backpropagated gradient error along both spatial and temporal dimensions. 

\textbf{SDBP} only retains the temporal propagation in the forward pass but omits its impact on the backward pass. This prevents the gradient from propagating across time, thereby limiting the learning of temporal dependency. The modified gradient error is:
\begin{align}
\boldsymbol{\hat{\delta}}^l[t] &= \begin{cases} 
\frac{\partial \mathcal{L}}{\partial \boldsymbol{u}^L[t]},  &l\!=\!L,\\[1.5ex] 
\boldsymbol{\hat{\delta}}^{l+1}[t]  \frac{\partial \boldsymbol{u}^{l+1}[t]}{\partial \boldsymbol{u}^{l}[t]},  &l\!<\!L.
\end{cases}
\label{eq:tlg2}
\end{align}

\textbf{NoTD} removes temporal propagation in both forward and backward passes, processing each time step independently. The resulting neuronal dynamic is simplified to:
\begin{align}
\boldsymbol{u}^{l}[t] &= \boldsymbol{W}^{l}  \boldsymbol{s}^{l-1}[t], \label{eq:notd_forward}
\end{align}
which excludes the update of $\boldsymbol{u}^{l}[t]$ compared to Eq.~\eqref{eq:mem_update}, rendering it unable to integrate any temporal information across time. The gradient in NoTD is the same as Eq. \eqref{eq:tlg2} in SDBP. 

By training an SNN using the three algorithms described and comparing their performance gaps, we can quantify the effectiveness of a specific task in assessing the temporal processing capacity of the SNN:
\begin{enumerate}
  \item A similar performance between STBP and SDBP indicates that the task fails to leverage temporal propagation in the backward pass to establish meaningful temporal dependencies, and, therefore, is ineffective in evaluating temporal processing capacity.
  \item A comparable performance between NoTD and STBP implies that the task can be solved without leveraging any temporal information, making it unsuitable as a benchmark for temporal processing. 
  \item In contrast, significant performance degradation in SDBP and NoTD compared to STBP indicates that the task contains rich temporal information and can effectively evaluate temporal processing capacity. 
\end{enumerate}

\begin{table*}[h!]
\centering
\caption{Validating the effectiveness of NSA in evaluating the temporal processing capacity of SNNs.}
\renewcommand{\arraystretch}{1.1}
\resizebox{0.84\textwidth}{!}{
\begin{tabular}{@{}l|c|c|c|c|c|c|c@{}}
\toprule
\textbf{Method} & \makecell[c]{\textbf{AL}\\Acc. (\%) $\uparrow$ } & \makecell[c]{\textbf{HAR}\\Acc. (\%) $\uparrow$} & \makecell[c]{\textbf{EEG-MI}\\Acc. (\%) $\uparrow$}  & \makecell[c]{\textbf{SSL}\\Acc. (\%) $\uparrow$} & \makecell[c]{\textbf{ALR}\\Acc. (\%) $\uparrow$} & \makecell[c]{\textbf{AD}\\\ SI-SNR (dB) $\uparrow$}& \makecell[c]{\textbf{ASR}\\CER (\%) $\downarrow$} \\ \midrule
STBP & \textbf{63.52} & \textbf{81.27}  & \textbf{65.20} & \textbf{8.88} & \textbf{17.83} &  \textbf{11.47}& \textbf{20.70} \\
SDBP & 58.52 (-5.00) & 76.29 (-4.98)  & 57.48 (-7.72)  & 5.79 (-3.09) & 2.47 (-15.36) & 10.18 (-1.29) &26.30 (+5.60)  \\
NoTD & 53.34 (-10.18) & 67.75 (-13.52)  & 52.48 (-12.72)  & 3.45 (-5.43) & 2.01 (-15.82) & 9.47 (-2.00)  &27.10 (+6.40)  \\ \bottomrule
\end{tabular}}
\label{tab:res_stp}
\end{table*}

\begin{table*}[h!]
\centering
\renewcommand{\arraystretch}{1.1}
\caption{Results of NSA benchmark for different spiking neuron models with SFNN or SRNN architectures.}
\resizebox{0.96\textwidth}{!}{
\begin{threeparttable}
\begin{tabular}{@{}l|l|ccccccc|c@{}}
\toprule
 \multirow{2}{*}{\textbf{Architecture}}& \multirow{2}{*}{\makecell[c]{\textbf{Neuron} \textbf{model}}} &  \textbf{AL} & \textbf{HAR} & \textbf{EEG-MI} & \textbf{SSL} & \textbf{ALR} & \textbf{AD} & \textbf{ASR} & \multirow{2}{*}{\makecell[c]{\textbf{Average}\\\textbf{rank} $\downarrow$}}\\ 
& &  Acc. (\%) $\uparrow$ & Acc. (\%) $\uparrow$& Acc. (\%) $\uparrow$ & Acc. (\%) $\uparrow$ & Acc. (\%) $\uparrow$ & SI-SNR (dB) $\uparrow$ & CER (\%) $\downarrow$ \\ \midrule
  \multirow{5}{*}{SFNN} & LIF  & 54.40 & 81.27 & 65.20 & 8.88 & 17.83 &11.47 & 20.70 & 6.6 \\
  &CE-LIF  & 59.58 & 79.71 & 70.56 & 12.21 &48.32 & -\tnote{*} &17.70 & 5.3\\
  &LTC & 77.40 & 80.97  & 64.42 & 10.33 &48.93 & \textbf{14.36} &{15.90} & 4.3 \\
  &SPSN  & 72.02 & \underline{86.73} & \textbf{76.46} & \underline{20.30} &45.73 &13.00 & 18.50 & 3.7 \\
  &PMSN  & \textbf{87.42} & \textbf{88.31} & \underline{75.00} & \textbf{75.35} &\textbf{57.43} & \underline{14.35} & 17.70 & \textbf{1.9}\\ 
  \midrule
  \multirow{3}{*}{SRNN}  & LIF  & 56.50 & 77.32 & 64.68 & 6.17 & 34.71& 9.36 &\underline{15.70} & 6.4\\
  & CE-LIF & 60.62 & 80.83 & {74.35} & 12.72 &51.64 & -\tnote{*} & \textbf{15.30} & 3.7 \\
  & LTC & \underline{79.04} & 82.03 & 69.16 & 16.37 & \underline{56.64}& 14.29 & 16.30 & \underline{3.1}\\ 
 \bottomrule
\end{tabular}
\begin{tablenotes}\footnotesize\item[*] These models are not applicable due to the inconsistent sequence lengths between the training and test phases.
\end{tablenotes}\end{threeparttable}}
\end{table*}

\begin{table*}[h!]
\centering
\renewcommand{\arraystretch}{1.1}
\caption{Results of NSA benchmark for different neural architectures using LIF neurons.}
\resizebox{0.96\textwidth}{!}{
\begin{threeparttable}
\begin{tabular}{@{}l|ccccccc|c@{}}
\toprule
  \multirow{2}{*}{\makecell[c]{\textbf{Architecture}}} &  \textbf{AL} & \textbf{HAR} & \textbf{EEG-MI} & \textbf{SSL} & \textbf{ALR} & \textbf{AD} & \textbf{ASR} & \multirow{2}{*}{\makecell[c]{\textbf{Average}\\\textbf{rank} $\downarrow$}}\\ 
 &  Acc. (\%) $\uparrow$ & Acc. (\%) $\uparrow$& Acc. (\%) $\uparrow$ & Acc. (\%) $\uparrow$ & Acc. (\%) $\uparrow$ & SI-SNR (dB) $\uparrow$ & CER (\%) $\downarrow$ \\ \midrule
 SFNN  & 54.40 & 81.27 & 65.20 & 8.88 & 17.83 &11.47 & 20.70 & 5.9 \\
 SRNN  & 56.50 & 77.32 & 64.68 & 6.17 & 34.71& 9.36 &15.70 & 5.9\\
  GSN & 67.22 & 82.31 & 68.85 & 10.34 &21.17 & \textbf{14.45} & \textbf{14.30} & 3.6 \\
Spiking TCN   & 69.88 & 82.41 & 75.58 & 40.60 & \textbf{47.14} & 12.77& 17.10 & 3.1\\
Spike-Driven Transformer  & 58.00 & 71.07 & 69.12 & 5.75 & 39.62 & 9.86 & 36.80 & 5.7\\
Binary S4D  & \textbf{81.44} & \textbf{89.37} & \textbf{78.61} & \underline{79.34} & \underline{44.80} &\underline{14.21} &15.20 &\textbf{1.7}\\
GSU   & \underline{80.42} & \underline{89.16} & \underline{77.43} & \textbf{82.39} &{41.35} &14.05 & \underline{14.40} &\underline{2.1}\\ \bottomrule
\end{tabular}\end{threeparttable}}
\end{table*}

\section{Results}
In this section, we first employ STP to validate the effectiveness of the existing and the proposed NSA benchmarks in evaluating the temporal processing capacity of SNNs. Following this, we conduct a comprehensive comparison of recently proposed spiking neuron models and neural architectures using NSA, thereby illuminating the current status of SNNs in temporal processing. Additionally, we incorporate deployment-critical efficiency metrics to further benchmark these models, including training speed, memory usage, and energy consumption, to assess their practicality for real-world applications. For all Tables, \textbf{bold values} indicate the best performance, and \underline{underlined values} represent the second best. Details on experimental configurations and hyperparameters are provided in Supplementary Materials. 

\subsection{Benchmark Effectiveness Validation}
\label{sec:effectivness}
We first apply the proposed STP tool to evaluate 12 commonly used benchmarks in the neuromorphic community (see Supplementary Materials for details). Our results indicate comparable performance between STBP and SDBP, suggesting that their effectiveness in assessing the temporal processing capacities of SNNs is limited. Subsequently, we perform the same study on the proposed NSA benchmark to validate its effectiveness. As shown in Table~\ref{tab:res_stp}, both SDBP and NoTD exhibit substantial performance degradation compared to STBP across all tasks in NSA. This suggests that the seven selected tasks contain essential temporal dependencies that must be effectively captured to attain high performance. Consequently, the proposed NSA serves as a more effective benchmark for neuromorphic temporal processing.

\subsection{Performance Benchmarking of SNN Models}
\label{sec:performance}

To elucidate the current status of SNN models in temporal processing, we further evaluate five spiking neuron models notable for their enhanced temporal processing capacities: LIF, Context Embedding LIF (CE-LIF) \cite{chen2023unleashing}, Liquid Time-Constant (LTC) \cite{yin2023accurate}, Sliding Parallel Spiking Neuron (SPSN) \cite{fang2024parallel}, and Parallel Multicompartment Spiking Neuron (PMSN) \cite{chen2024pmsn}. In addition to the commonly used feedforward (SFNN) and recurrent (SRNN) networks \cite{bellec2018long}, we also benchmark five advanced neural architectures that excel in temporal processing, including Gated Spiking Neuron (GSN) \cite{hao2024towards}, Spiking Temporal Convolution Network (Spiking TCN) \cite{bai2018empirical}, Spike-Driven Transformer \cite{yao2024spike}, Binary S4D \cite{stan2024learning}, and Gated Spiking Unit (GSU) \cite{stan2024learning}.
%\footnote{To avoid confusion, GSU in \cite{stan2024learning} is renamed as GSU-SSM in this paper for differentiation.}  
To ensure a fair comparison, all models are configured with a comparable number of trainable parameters for each task. For architectures like Spiking TCN and Spike-Driven Transformer, which are originally designed to repeat each time step $D$ times to enhance their model representation power, we set $D=1$ to facilitate fair comparisons with other models.

Our experimental results demonstrate that the performance of spiking neuron models varies significantly across tasks. LTC shows only marginal improvements in noise-intensive tasks such as HAR, EEG-MI, and SSL, attributable to its input-dependent mechanism, which is highly susceptible to noise accumulation over time. In contrast, PMSN and SPSN neurons exhibit stronger noise robustness in these tasks. Additionally, PMSN and LTC excel in AL and ALR tasks requiring long-term memory, showcasing superior temporal retention capacities over extended periods. Furthermore, LTC and PMSN demonstrate high expressiveness in the AD task, while CE-LIF with recurrent connections stands out in the ASR task, exhibiting remarkable generalizability. Overall, PMSN emerges as the top-performing spiking neuron model on NSA, striking a balance between robustness, temporal dependency establishment, and generalizability across tasks.

For neural architectures, Binary S4D and its variant GSU exhibit strong noise robustness in HAR, EEG-MI, and SSL tasks, outperforming other architectures in handling noisy inputs. In contrast, the GSN model achieves notable gains in AD and ASR tasks, which require high model expressiveness and generalizability. This result suggests the high effectiveness of GSN in capturing diverse, fine-grained temporal patterns. Conversely, SRNNs and Spike-Driven Transformer fall short in performance compared to other models. We noticed that SRNNs often suffer from training instability and convergence issues in our experiments, resulting in comparable or worse performance than SFNNs. The Spike-Driven Transformer struggles to establish temporal dependencies effectively due to its binary activations (i.e., $D=1$), which significantly constrains its expressiveness. Meanwhile, this architecture is typically designed to work with large datasets and networks, the moderate dataset size of NSA and the small parameter count in our evaluation may constrain the full exploitation of its architectural advantages.
%This model is originally designed to repeat each time step $D$ times to expand representation resolution. However, we set $D=1$ for fair comparisons with other models, leading to 

\subsection{Efficiency Benchmarking of SNN Models}
\label{sec:efficiency}
We further report the efficiency metrics of the evaluated SNN models on the AL task, including training speed, GPU memory usage, and energy efficiency, which are critical factors for practical implementation. To ensure a fair comparison, all evaluations are conducted using a batch size of 256 and all evaluated SNN models are configured with a uniform network dimension, comprising two hidden layers with 256 channels each.
Training speed and memory costs are evaluated with sequence lengths of \{200, 400, 800\}, while energy efficiency is assessed with a sequence length of 400. The comparative results are summarized in Table \ref{tab:efficiency}.

\begin{table*}[]
    \caption{Comparison of SNN models in terms of the training speed, memory consumption, and energy efficiency. Values in brackets indicate ratios relative to the baseline LIF-SFNN model.}
    %\vspace{-0.1mm}
    \renewcommand{\arraystretch}{1.3}
    \centering
    \resizebox{0.99\textwidth}{!}{
\begin{tabular}{@{}l|l|ccc|ccc|ccc@{}}
\toprule
\multirow{2}{*}{\textbf{Architecture}} & \multirow{2}{*}{\makecell[c]{\textbf{Neuron}\\ \textbf{model}}} 
& \multicolumn{3}{c|}{\textbf{Training speed} (k step/s) $\uparrow$} 
& \multicolumn{3}{c|}{\textbf{Memory consumption} (GB) $\downarrow$} 
& \multicolumn{3}{c}{\makecell[c] {\textbf{Energy} \textbf{efficiency}}} \\
         & & 200 & 400 & 800  & 200 & 400 & 800 & \textbf{ACs} (k) $\downarrow$ & \textbf{MACs} (k) $\downarrow$ & \textbf{Empirical cost} (nJ) $\downarrow$ \\
\midrule
\multirow{5}{*}{SFNN} 
& LIF & 1.91 (1.0) & 1.91 (1.0) & 2.04 (1.0) & \textbf{0.49 (1.0)} & \textbf{0.98 (1.0)} & \textbf{1.96 (1.0)} & 3.72 (1.0) & 0.26 (1.0) & \textbf{4.52 (1.0)} \\
& CE-LIF & 1.25 (0.7) & 1.31 (0.7) & 1.37 (0.7) & 0.59 (1.2) & 1.18 (1.2) & 2.35 (1.2) & 3.22 (0.9) & 0.77 (3.0) & \underline{6.43 (1.4)} \\
& LTC & 0.63 (0.3) & 0.66 (0.3) & 0.68 (0.3) & 1.43 (2.9) & 2.84 (2.9) & 5.68 (2.9) & \textbf{0.93 (0.3)} & 262.91 (1,011)& 1,210.23 (268) \\
& SPSN & 5.04 (2.6) & 4.14 (2.2) & 3.21 (1.6) & \underline{0.50 (1.0)} & \underline{1.01 (1.0)} & \underline{2.01 (1.0)} & \underline{1.56 (0.4)} & 32.77 (126) & 152.14 (33) \\
& PMSN & \underline{6.34 (3.3)} & \underline{6.31 (3.3)} & \underline{6.24 (3.1)} & 1.23 (2.5) & 2.46 (2.5) & 4.92 (2.5) & 10.27 (2.8) & 4.66 (18) & 30.68 (6.8) \\
\midrule
\multirow{3}{*}{SRNN} 
& LIF & 1.19 (0.6) & 1.25 (0.7) & 1.30 (0.6) & 0.54 (1.1) & 1.08 (1.1) & 2.16 (1.1) & 22.91 (6.2) & 0.26 (1.0) & 21.79 (4.8) \\
& CE-LIF & 0.92 (0.5) & 0.98 (0.5) & 1.00 (0.5)  & 0.64 (1.3) & 1.28 (1.3) & 2.55 (1.3) & 8.26 (2.2) & 0.77 (3.0) & 10.96 (2.4) \\
& LTC & 0.55 (0.3) & 0.57 (0.3) & 0.57 (0.3)  & 1.43 (2.9) & 2.84 (2.9) & 5.68 (2.9) & 2.12 (0.6) & 262.91 (1,011)   &1,211.30 (268) \\
\midrule
GSN & & 0.88 (0.5) & 0.92 (0.5) & 0.92 (0.5)  & 0.64 (1.3) & 1.28 (1.3) & 2.55 (1.3) & 27.33 (7.3) & 1.28 (4.9) & 30.48 (6.7) \\
Spiking TCN & & 4.92 (2.6) & 4.99 (2.6) & 5.01 (2.5)  & 0.66 (1.4) & 1.30 (1.3) & 2.58 (1.3) & 50.49 (11) & \textbf{0.00 (0.0)} & 56.10 (15) \\
Spike-Driven Transformer & LIF & 1.86 (1.0) & 1.99 (1.0) & 2.06 (1.0)  & 3.07 (6.3) & 6.10 (6.2) & 12.16 (6.2) & 214.70 (58) & \textbf{0.00 (0.0)}  & 193.22 (42) \\
Binary S4D & & \textbf{7.04 (3.7)} & \textbf{7.07 (3.7)} & \textbf{7.33 (3.6)} & 1.26 (2.6) & 2.51 (2.6) & 5.00 (2.6) &44.49 (12) & 5.43 (21) & 65.01 (14) \\
GSU & & 5.93 (3.1) & 5.91 (3.1) & 5.92 (2.9)  & 1.48 (3.0) & 2.95 (3.0) & 5.88 (3.0) & 6.37 (1.7) & 4.92 (19)  & 28.36 (6.3) \\
\bottomrule
\end{tabular}}
    \label{tab:efficiency}
\end{table*}

\subsubsection{Training speed}
%The training speed of SNN models presents a significant bottleneck for processing long temporal sequences. 
Our evaluation results on NSA exhibit notable training speed differences between serial and parallel models. Serial models, including LIF, CE-LIF, LTC, and GSN, process temporal data sequentially, maintaining consistent but inherently slow training speeds per time step, regardless of sequence lengths. Consequently, their total training time scales linearly with the sequence length, leading to high computational costs for processing long sequences. This limitation becomes even more pronounced when involving complex neuronal dynamics. For instance, LIF achieves the fastest training speed among serial models, whereas LTC achieves the slowest, highlighting that inefficient training in serial models poses a significant bottleneck for long sequence processing.

Parallel models, on the contrary, demonstrate high computational efficiency by processing data in multiple time steps simultaneously. Our results show that recently proposed parallel models, including SPSN, PMSN, Spiking TCN, binary S4D, and GSU, achieve approximately $3~\times$ speedup compared to serial models. However, the Spike-Driven Transformer fails to achieve noticeable speedups over the LIF model due to its high computational complexity. 
Additionally, the training speed of parallel models varies with sequence length, reflecting differences in their underlying parallelization strategies. 
% Specifically, SPSN exhibits an increment in computation time per time step, leading to exponentially growing total training time. This can be attributed to the expansion of its receptive kernel size in conjunction with the sequence length, allowing for the capture of longer temporal dependencies. For other parallel models, the total time increases proportionally with the sequence length. Despite this proportionality, their reduced per-time step overhead compared to serial models makes them more attractive for tasks requiring long sequence processing.
Specifically, SPSN exhibits a decline in training speed as the sequence length increases, resulting in an exponentially growing total training time. This slowdown arises from the design of its receptive kernel, which expands proportionally with sequence length to capture long-range temporal dependencies. Consequently, the neuron dynamics incur a quadratic time complexity of $\mathcal{O}(L^2)$. In contrast, other parallel models sustain consistent training speeds across varying sequence lengths, owing to their more efficient linear-time complexity of $\mathcal{O}(L)$.

\subsubsection{Memory usage}
Despite enhanced temporal processing capacity offered by many recent SNN models, our finding reveals that these advancements come with a significant increase in memory consumption. Specifically, 
while parallel computing models accelerate training speeds, they typically incur substantially higher memory usage, trading off between space and time. This inefficiency arises from their need to store additional states to capture temporal dependencies in parallel. 
Among these parallel models, the Spike-Driven Transformer stands out as the most memory-intensive due to the expanded embedding space involved in the self-attention mechanism. In contrast, PMSN, S4D, and GSU consume comparatively less memory, benefiting from their compact representation of model states. However, they still exceed the memory requirements of serial approaches due to the additional storage needed for their parallelized operations. SPSN employs a 1-D temporal convolution kernel to capture local temporal features without the need to buffer extra transient states, making it a highly memory-efficient parallel architecture comparable to the LIF model. In contrast, serial models generally consume less memory as they store fewer transient states across time. The only exception is LTC, whose poor memory efficiency stems from its complex computation graph involved in gating computations. Notably, memory consumption for all evaluated models scales linearly with sequence length, regardless of structure or computational strategy. It highlights SNNs' memory efficiency in handling extended temporal sequences, as each time step contributes a fixed and predictable amount of memory without exponential growth.

\subsubsection{Energy efficiency}
The high energy efficiency of SNNs promises to enable efficient temporal processing at the edge. To present a comprehensive assessment of the energy efficiency of advanced SNN models, we conduct a quantitative analysis of the number of Multiply-Accumulate (MAC) and Accumulate (AC) operations per inference time step and per sample. Additionally, we derive the average empirical energy consumption for each model based on the experimental data. Detailed calculations of energy cost can be found in the Supplementary Materials.

The results presented in Table \ref{tab:efficiency} suggest that most advanced SNN models improve temporal processing capacity at the cost of increased energy consumption. Among the evaluated models, CE-LIF, PMSN, and GSU stand out for achieving a favorable trade-off between performance and energy efficiency, making them more suitable for deployment in resource-constrained edge systems.
It is noteworthy that LTC presents two orders of magnitude higher energy consumption than standard LIF models, which can be attributed to the hundreds of thousands of MAC operations required for computing its temporal dynamics. Such high energy costs comparable to traditional ANN methods severely limit its practicality for energy-constrained systems.
These observations highlight the critical role of efficiency evaluation in model design. To unleash the full potential of neuromorphic systems in temporal processing, we argue that the development of SNN models must prioritize not only the enhancement of task accuracy but also the maintenance of ultra-low energy consumption, thereby ensuring their competitiveness in real-world applications.

\section{Related Works: Current SNN Benchmarking Practices}

{Existing benchmarks commonly used for evaluating SNNs can be divided into four categories, each with its own limitations that hinder their ability to effectively assess the temporal processing capacity of SNNs.}
The first category includes static image recognition tasks~\cite{lecun1998gradient,krizhevsky2009learning}, where identical images are repeated along the time axis, lacking any meaningful temporal dynamics. 
The second category comprises event-based visual classification tasks recorded by DVS cameras~\cite{li2017cifar10,amir2017low,zhou2024enhancing,wang2022event,wang2024har}. 
While these datasets impose artificial saccadic motion on static images or capture simple moving objects, their limited temporal dynamics result in performance that primarily emphasizes spatial pattern recognition rather than the establishment of long-range temporal dependencies. The third category involves keyword spotting tasks, which encompass both frame-based~\cite{warden2018speech} and spike-based~\cite{cramer2020heidelberg} audio inputs. While these datasets contain richer temporal dynamics, effective decisions can often be made by integrating only short-term temporal features, making these datasets insufficiently challenging to evaluate the temporal processing capacity of SNNs. More recently, several preliminary efforts have been made to apply SNNs to long-term language modeling tasks~\cite{tay2020long}. Despite the complex temporal dependencies inherent in these tasks, the high training costs associated with these models render them unsuitable for evaluating many existing SNN approaches. Furthermore, addressing such tasks typically necessitates models with a substantial number of parameters, which are not feasible for deployment on current neuromorphic hardware, thereby limiting their utility as benchmarks for SNNs at this stage of development.

\section{Discussion and Conclusion }
In this work, we present NSA, an effective, versatile, and application-oriented benchmark designed to comprehensively evaluate the temporal processing capacities of SNNs across diverse application scenarios. To ensure rigorous and reliable assessment, we integrate a temporal dependency analysis tool, STP, into NSA to quantify the effectiveness of the benchmark.
Our comparative analysis underscores both the progress and the challenges in neuromorphic temporal
processing. While advanced spiking neuron models and neural architectures demonstrate remarkable improvements in task performance, many of them struggle to maintain efficiency in training speed, memory usage, and energy consumption, which are crucial constraints for real-world applications. {Our findings underscore} the urgent need to develop effective SNN models capable of robustly processing temporal data while maintaining high energy efficiency.
While this paper provides a limited evaluation of SNN approaches due to time and resource constraints, we encourage the community to expand the scope of evaluations using NSA. We envision NSA as an effective and adaptive temporal benchmarking framework capable of addressing the evolving needs of the community. We hope this benchmark will inspire further advancements in neuromorphic temporal processing research, thereby paving the way for more capable, robust, and efficient neuromorphic solutions for real-world applications.

\section*{Acknowledgments}
This work was partially supported by the National Natural Science Foundation of China (Grant No. 62306259 and U21A20512), the Research Grants Council of the Hong Kong SAR (Grant No. PolyU25216423, PolyU11211521, PolyU15218622, PolyU15215623, and C5052-23G), and The Hong Kong Polytechnic University (Project IDs: P0043563, P0046094).

% \bibliographystyle{named}
% \bibliography{ijcai25}

\begin{thebibliography}{}

\bibitem[\protect\citeauthoryear{Amir \bgroup \em et al.\egroup }{2017}]{amir2017low}
Arnon Amir, Brian Taba, David Berg, Timothy Melano, Jeffrey McKinstry, Carmelo Di~Nolfo, Tapan Nayak, Alexander Andreopoulos, Guillaume Garreau, Marcela Mendoza, et~al.
\newblock A low power, fully event-based gesture recognition system.
\newblock In {\em Proceedings of the IEEE Conference on Computer Vision and Pattern Recognition}, pages 7243--7252, 2017.

\bibitem[\protect\citeauthoryear{Bai \bgroup \em et al.\egroup }{2018}]{bai2018empirical}
Shaojie Bai, J~Zico Kolter, and Vladlen Koltun.
\newblock An empirical evaluation of generic convolutional and recurrent networks for sequence modeling.
\newblock {\em arXiv preprint arXiv:1803.01271}, 2018.

\bibitem[\protect\citeauthoryear{Bellec \bgroup \em et al.\egroup }{2018}]{bellec2018long}
Guillaume Bellec, Darjan Salaj, Anand Subramoney, Robert Legenstein, and Wolfgang Maass.
\newblock Long short-term memory and learning-to-learn in networks of spiking neurons.
\newblock {\em Advances in neural information processing systems}, 31, 2018.

\bibitem[\protect\citeauthoryear{Bu \bgroup \em et al.\egroup }{2017}]{bu2017aishell}
Hui Bu, Jiayu Du, Xingyu Na, Bengu Wu, and Hao Zheng.
\newblock Aishell-1: An open-source mandarin speech corpus and a speech recognition baseline.
\newblock In {\em 2017 20th conference of the oriental chapter of the international coordinating committee on speech databases and speech I/O systems and assessment (O-COCOSDA)}, pages 1--5. IEEE, 2017.

\bibitem[\protect\citeauthoryear{Burkitt}{2006}]{burkitt2006review}
Anthony~N Burkitt.
\newblock A review of the integrate-and-fire neuron model: I. homogeneous synaptic input.
\newblock {\em Biological cybernetics}, 95:1--19, 2006.

\bibitem[\protect\citeauthoryear{Chen \bgroup \em et al.\egroup }{2023}]{chen2023unleashing}
Xinyi Chen, Jibin Wu, Huajin Tang, Qinyuan Ren, and Kay~Chen Tan.
\newblock Unleashing the potential of spiking neural networks for sequential modeling with contextual embedding.
\newblock {\em arXiv preprint arXiv:2308.15150}, 2023.

\bibitem[\protect\citeauthoryear{Chen \bgroup \em et al.\egroup }{2024}]{chen2024pmsn}
Xinyi Chen, Jibin Wu, Chenxiang Ma, Yinsong Yan, Yujie Wu, and Kay~Chen Tan.
\newblock Pmsn: A parallel multi-compartment spiking neuron for multi-scale temporal processing.
\newblock {\em arXiv preprint arXiv:2408.14917}, 2024.

\bibitem[\protect\citeauthoryear{Cramer \bgroup \em et al.\egroup }{2020}]{cramer2020heidelberg}
Benjamin Cramer, Yannik Stradmann, Johannes Schemmel, and Friedemann Zenke.
\newblock The heidelberg spiking data sets for the systematic evaluation of spiking neural networks.
\newblock {\em IEEE Transactions on Neural Networks and Learning Systems}, 33(7):2744--2757, 2020.

\bibitem[\protect\citeauthoryear{Davies \bgroup \em et al.\egroup }{2018}]{davies2018loihi}
Mike Davies, Narayan Srinivasa, Tsung-Han Lin, Gautham Chinya, Yongqiang Cao, Sri~Harsha Choday, Georgios Dimou, Prasad Joshi, Nabil Imam, Shweta Jain, et~al.
\newblock Loihi: A neuromorphic manycore processor with on-chip learning.
\newblock {\em IEEE Micro}, 38(1):82--99, 2018.

\bibitem[\protect\citeauthoryear{Deng \bgroup \em et al.\egroup }{2022}]{deng2022temporal}
Shikuang Deng, Yuhang Li, Shanghang Zhang, and Shi Gu.
\newblock Temporal efficient training of spiking neural network via gradient re-weighting.
\newblock In {\em International Conference on Learning Representations}, pages 1--14, 2022.

\bibitem[\protect\citeauthoryear{Fang \bgroup \em et al.\egroup }{2023}]{fang2024parallel}
Wei Fang, Zhaofei Yu, Zhaokun Zhou, Ding Chen, Yanqi Chen, Zhengyu Ma, Timoth\'{e}e Masquelier, and Yonghong Tian.
\newblock Parallel spiking neurons with high efficiency and ability to learn long-term dependencies.
\newblock In {\em Advances in Neural Information Processing Systems}, volume~36, pages 53674--53687, 2023.

\bibitem[\protect\citeauthoryear{Hao \bgroup \em et al.\egroup }{2024}]{hao2024towards}
Xiang Hao, Chenxiang Ma, Qu~Yang, Jibin Wu, and Kay~Chen Tan.
\newblock Towards ultra-low-power neuromorphic speech enhancement with spiking-fullsubnet.
\newblock {\em arXiv preprint arXiv:2410.04785}, 2024.

\bibitem[\protect\citeauthoryear{He \bgroup \em et al.\egroup }{2024}]{he2024network}
Linxuan He, Yunhui Xu, Weihua He, Yihan Lin, Yang Tian, Yujie Wu, Wenhui Wang, Ziyang Zhang, Junwei Han, Yonghong Tian, et~al.
\newblock Network model with internal complexity bridges artificial intelligence and neuroscience.
\newblock {\em Nature Computational Science}, 4(8):584--599, 2024.

\bibitem[\protect\citeauthoryear{Horowitz}{2014}]{CMOS}
Mark Horowitz.
\newblock 1.1 computing's energy problem (and what we can do about it).
\newblock In {\em 2014 IEEE International Solid-State Circuits Conference Digest of Technical Papers (ISSCC)}, pages 10--14. IEEE, 2014.

\bibitem[\protect\citeauthoryear{Hosseini \bgroup \em et al.\egroup }{2020}]{hosseini2020review}
Mohammad-Parsa Hosseini, Amin Hosseini, and Kiarash Ahi.
\newblock A review on machine learning for eeg signal processing in bioengineering.
\newblock {\em IEEE reviews in biomedical engineering}, 14:204--218, 2020.

\bibitem[\protect\citeauthoryear{Krizhevsky and Hinton}{2009}]{krizhevsky2009learning}
Alex Krizhevsky and Geoffrey Hinton.
\newblock Learning multiple layers of features from tiny images.
\newblock Technical Report~0, University of Toronto, Toronto, Ontario, 2009.

\bibitem[\protect\citeauthoryear{LeCun \bgroup \em et al.\egroup }{1998}]{lecun1998gradient}
Yann LeCun, L{\'e}on Bottou, Yoshua Bengio, and Patrick Haffner.
\newblock Gradient-based learning applied to document recognition.
\newblock {\em Proceedings of the IEEE}, 86(11):2278--2324, 1998.

\bibitem[\protect\citeauthoryear{Lee \bgroup \em et al.\egroup }{2019}]{lee2019}
Min-Ho Lee, O-Yeon Kwon, Yong-Jeong Kim, Hong-Kyung Kim, Young-Eun Lee, John Williamson, Siamac Fazli, and Seong-Whan Lee.
\newblock Eeg dataset and openbmi toolbox for three bci paradigms: an investigation into bci illiteracy.
\newblock {\em GigaScience}, 8(5):giz002, 01 2019.

\bibitem[\protect\citeauthoryear{Li \bgroup \em et al.\egroup }{2017}]{li2017cifar10}
Hongmin Li, Hanchao Liu, Xiangyang Ji, Guoqi Li, and Luping Shi.
\newblock Cifar10-dvs: an event-stream dataset for object classification.
\newblock {\em Frontiers in Neuroscience}, 11:244131, 2017.

\bibitem[\protect\citeauthoryear{Ma \bgroup \em et al.\egroup }{2024}]{ma2024darwin3}
De~Ma, Xiaofei Jin, Shichun Sun, Yitao Li, Xundong Wu, Youneng Hu, Fangchao Yang, Huajin Tang, Xiaolei Zhu, Peng Lin, et~al.
\newblock Darwin3: a large-scale neuromorphic chip with a novel isa and on-chip learning.
\newblock {\em National Science Review}, 11(5):nwae102, 2024.

\bibitem[\protect\citeauthoryear{Maass}{1997}]{maass1997networks}
Wolfgang Maass.
\newblock Networks of spiking neurons: the third generation of neural network models.
\newblock {\em Neural Networks}, 10(9):1659--1671, 1997.

\bibitem[\protect\citeauthoryear{Orchard \bgroup \em et al.\egroup }{2015}]{orchard2015converting}
Garrick Orchard, Ajinkya Jayawant, Gregory~K Cohen, and Nitish Thakor.
\newblock Converting static image datasets to spiking neuromorphic datasets using saccades.
\newblock {\em Frontiers in neuroscience}, 9:437, 2015.

\bibitem[\protect\citeauthoryear{Pei \bgroup \em et al.\egroup }{2019}]{pei2019towards}
Jing Pei, Lei Deng, Sen Song, Mingguo Zhao, Youhui Zhang, Shuang Wu, Guanrui Wang, Zhe Zou, Zhenzhi Wu, Wei He, et~al.
\newblock Towards artificial general intelligence with hybrid tianjic chip architecture.
\newblock {\em Nature}, 572(7767):106--111, 2019.

\bibitem[\protect\citeauthoryear{Qian \bgroup \em et al.\egroup }{2021}]{qian2021sloclas}
Xinyuan Qian, Bidisha Sharma, Amine~El Abridi, and Haizhou Li.
\newblock Sloclas: A database for joint sound localization and classification.
\newblock {\em arXiv preprint arXiv:2108.02539}, 2021.

\bibitem[\protect\citeauthoryear{Stan and Rhodes}{2024}]{stan2024learning}
Matei-Ioan Stan and Oliver Rhodes.
\newblock Learning long sequences in spiking neural networks.
\newblock {\em Scientific Reports}, 14(1):21957, 2024.

\bibitem[\protect\citeauthoryear{Sun \bgroup \em et al.\egroup }{2024}]{sun2024delayed}
Pengfei Sun, Jibin Wu, Malu Zhang, Paul Devos, and Dick Botteldooren.
\newblock Delayed memory unit: Modeling temporal dependency through delay gate.
\newblock {\em IEEE Transactions on Neural Networks and Learning Systems}, 2024.

\bibitem[\protect\citeauthoryear{Tan \bgroup \em et al.\egroup }{2022}]{tan2022multi}
Ganchao Tan, Yang Wang, Han Han, Yang Cao, Feng Wu, and Zheng-Jun Zha.
\newblock Multi-grained spatio-temporal features perceived network for event-based lip-reading.
\newblock In {\em Proceedings of the IEEE/CVF Conference on Computer Vision and Pattern Recognition}, pages 20094--20103, 2022.

\bibitem[\protect\citeauthoryear{Tay \bgroup \em et al.\egroup }{2020}]{tay2020long}
Yi~Tay, Mostafa Dehghani, Samira Abnar, Yikang Shen, Dara Bahri, Philip Pham, Jinfeng Rao, Liu Yang, Sebastian Ruder, and Donald Metzler.
\newblock Long range arena: A benchmark for efficient transformers.
\newblock {\em arXiv preprint arXiv:2011.04006}, 2020.

\bibitem[\protect\citeauthoryear{Timcheck \bgroup \em et al.\egroup }{2023}]{timcheck2023intel}
Jonathan Timcheck, Sumit~Bam Shrestha, Daniel Ben~Dayan Rubin, Adam Kupryjanow, Garrick Orchard, Lukasz Pindor, Timothy Shea, and Mike Davies.
\newblock The intel neuromorphic dns challenge.
\newblock {\em Neuromorphic Computing and Engineering}, 3(3):034005, 2023.

\bibitem[\protect\citeauthoryear{Wang \bgroup \em et al.\egroup }{2022}]{wang2022event}
Yanxiang Wang, Xian Zhang, Yiran Shen, Bowen Du, Guangrong Zhao, Lizhen Cui, and Hongkai Wen.
\newblock Event-stream representation for human gaits identification using deep neural networks.
\newblock {\em IEEE Transactions on Pattern Analysis and Machine Intelligence}, 44(7):3436--3449, 2022.

\bibitem[\protect\citeauthoryear{Wang \bgroup \em et al.\egroup }{2024}]{wang2024har}
Xiao Wang, Zongzhen Wu, Bo~Jiang, Zhimin Bao, Lin Zhu, Guoqi Li, Yaowei Wang, and Yonghong Tian.
\newblock Hardvs: Revisiting human activity recognition with dynamic vision sensors.
\newblock {\em Proceedings of the AAAI Conference on Artificial Intelligence}, 38(6):5615--5623, Mar. 2024.

\bibitem[\protect\citeauthoryear{Warden}{2018}]{warden2018speech}
Pete Warden.
\newblock Speech commands: A dataset for limited-vocabulary speech recognition.
\newblock {\em arXiv preprint arXiv:1804.03209}, 2018.

\bibitem[\protect\citeauthoryear{Watanabe \bgroup \em et al.\egroup }{2018}]{watanabe2018espnet}
Shinji Watanabe, Takaaki Hori, Shigeki Karita, Tomoki Hayashi, Jiro Nishitoba, Yuya Unno, Nelson Enrique~Yalta Soplin, Jahn Heymann, Matthew Wiesner, Nanxin Chen, et~al.
\newblock Espnet: End-to-end speech processing toolkit.
\newblock {\em arXiv preprint arXiv:1804.00015}, 2018.

\bibitem[\protect\citeauthoryear{Weiss}{2019}]{weiss2019wisdm}
Gary~M Weiss.
\newblock Wisdm smartphone and smartwatch activity and biometrics dataset.
\newblock {\em UCI Machine Learning Repository: WISDM Smartphone and Smartwatch Activity and Biometrics Dataset Data Set}, 7:133190--133202, 2019.

\bibitem[\protect\citeauthoryear{Wu \bgroup \em et al.\egroup }{2018}]{wu2018spatio}
Yujie Wu, Lei Deng, Guoqi Li, Jun Zhu, and Luping Shi.
\newblock Spatio-temporal backpropagation for training high-performance spiking neural networks.
\newblock {\em Frontiers in Neuroscience}, 12:331, 2018.

\bibitem[\protect\citeauthoryear{Yang \bgroup \em et al.\egroup }{2024}]{yang2024vision}
Zheyu Yang, Taoyi Wang, Yihan Lin, Yuguo Chen, Hui Zeng, Jing Pei, Jiazheng Wang, Xue Liu, Yichun Zhou, Jianqiang Zhang, et~al.
\newblock A vision chip with complementary pathways for open-world sensing.
\newblock {\em Nature}, 629(8014):1027--1033, 2024.

\bibitem[\protect\citeauthoryear{Yao \bgroup \em et al.\egroup }{2024}]{yao2024spike}
Man Yao, Jiakui Hu, Zhaokun Zhou, Li~Yuan, Yonghong Tian, Bo~Xu, and Guoqi Li.
\newblock Spike-driven transformer.
\newblock {\em Advances in neural information processing systems}, 36, 2024.

\bibitem[\protect\citeauthoryear{Yin \bgroup \em et al.\egroup }{2021}]{yin2021accurate}
Bojian Yin, Federico Corradi, and Sander~M Boht{\'e}.
\newblock Accurate and efficient time-domain classification with adaptive spiking recurrent neural networks.
\newblock {\em Nature Machine Intelligence}, 3(10):905--913, 2021.

\bibitem[\protect\citeauthoryear{Yin \bgroup \em et al.\egroup }{2023}]{yin2023accurate}
Bojian Yin, Federico Corradi, and Sander~M Boht{\'e}.
\newblock Accurate online training of dynamical spiking neural networks through forward propagation through time.
\newblock {\em Nature Machine Intelligence}, 5(5):518--527, 2023.

\bibitem[\protect\citeauthoryear{Zhang \bgroup \em et al.\egroup }{2024}]{zhang2024tc}
Shimin Zhang, Qu~Yang, Chenxiang Ma, Jibin Wu, Haizhou Li, and Kay~Chen Tan.
\newblock Tc-lif: A two-compartment spiking neuron model for long-term sequential modelling.
\newblock In {\em Proceedings of the AAAI Conference on Artificial Intelligence}, volume~38, pages 16838--16847, 2024.

\bibitem[\protect\citeauthoryear{Zheng \bgroup \em et al.\egroup }{2024}]{zheng2024temporal}
Hanle Zheng, Zhong Zheng, Rui Hu, Bo~Xiao, Yujie Wu, Fangwen Yu, Xue Liu, Guoqi Li, and Lei Deng.
\newblock Temporal dendritic heterogeneity incorporated with spiking neural networks for learning multi-timescale dynamics.
\newblock {\em Nature Communications}, 15(1):277, 2024.

\bibitem[\protect\citeauthoryear{Zhou \bgroup \em et al.\egroup }{2024}]{zhou2024enhancing}
Shibo Zhou, Bo~Yang, Mengwen Yuan, Runhao Jiang, Rui Yan, Gang Pan, and Huajin Tang.
\newblock Enhancing snn-based spatio-temporal learning: A benchmark dataset and cross-modality attention model.
\newblock {\em Neural Networks}, 180:106677, 2024.

\end{thebibliography}

\newpage
\onecolumn
\begin{center}
    {\bf \LARGE Supplementary Materials}\\
    \vspace{0.2cm}
    { \Large Neuromorphic Sequential Arena: A Benchmark for Neuromorphic Temporal Processing}
    \vspace{0.2cm}
    
    {
Xinyi Chen$^{1}$,
Chenxiang Ma$^{1}$,
Yujie Wu$^{2}$,
Kay Chen Tan$^{1,3}$,
Jibin Wu$^{1,2,3}$}
\vspace{0.2cm}

{
$^1$Department of Data Science and Artificial Intelligence, The Hong Kong Polytechnic University\\
$^2$Department of Computing, The Hong Kong Polytechnic University\\
$^3$Research Center of Data Science and
Artificial Intelligence, The Hong Kong Polytechnic University\\
}
\end{center}
\rule[-0.5pt]{18.1cm}{0.06em}

\parskip=2pt

\renewcommand{\thesection}{S\arabic{section}}
\renewcommand{\thefigure}{S\arabic{figure}}
\renewcommand{\thetable}{S\arabic{table}}
\renewcommand{\theequation}{S\arabic{equation}}

\setcounter{section}{0}
\setcounter{figure}{0}
\setcounter{table}{0}
\setcounter{equation}{0}

\section*{Theoretical and Empirical Computation of Energy Cost}

In this section, we detail the derivation of energy consumption as presented in Table \ref{tab:efficiency}. Following standard practices for assessing energy costs of Spiking Neural Networks (SNNs), we quantify the number of 32-bit floating-point Multiply-Accumulate (MAC) operations and Accumulate (AC) operations involved in a single inference.

First, based on the temporal dynamics of each model, we derive theoretical energy consumption formulas, which are provided in Table \ref{tab:energy}. In these formulas, $E_{MAC}$ and $E_{AC}$ represent the energy required for each MAC and AC operation on neuromorphic hardware, respectively. $m$, $n$, and $k$ denote the input size, hidden size, and kernel size, respectively. $h$ is the hidden dimension of the feedforward module in Spike-Driven Transformer. The variable $Fr_i$ captures the spike frequency of a spike sequence $i$, with examples including $Fr_{in}$ and $Fr_{out}$, which represent the sparsity of input and output spike sequences for a given layer. For models with additional architectural components, such as the Spiking TCN, $Fr_{conv2}$ specifies the sparsity of input spikes in the second convolution layer. Similarly, in the Spike-Driven Transformer, $Fr_q$, $Fr_k$, $Fr_v$, $Fr_{attn}$, $Fr_{fc1}$, and $Fr_{fc2}$ denote the sparsity of LIF neuron populations in the self-attention and feedforward modules. For GSU, $Fr_{y}$ and $Fr_{w}$ represent the sparsity of ternary outputs and weights, respectively.

Subsequently, to derive these spike frequency values required for calculating the empirical energy consumption of the evaluated models, we perform inference on well-trained spiking models using the AL task with a sequence length of 400. All networks are configured with hidden dimensions of [256, 256], and the second hidden layer is consistently utilized for comparisons across evaluated models. During inference, we record the average spike frequency per time step, sample, and neuron for all spiking neuron populations involved in the computation.

Finally, based on data derived from a $45\ nm$ CMOS process \cite{CMOS}, the energy per operation is set to $E_{AC} = 0.9\ \text{pJ}$ for AC operations and $E_{MAC} = 4.6\ \text{pJ}$ for MAC operations. All these data are substituted into the theoretical energy cost formulas to compute the empirical energy cost, as outlined in the main text.

\begin{table}[h!]
\centering
\caption{Theoretical and empirical energy costs of different architectures and neuron models.}
\resizebox{0.8\textwidth}{!}{
\begin{tabular}{@{}l|l|c|c@{}}
\toprule
\textbf{Architecture} & \makecell[c]{\textbf{Neuron} \textbf{model}} & \textbf{Theoretical energy cost} & \makecell[c]{\textbf{Empirical} \textbf{energy cost} (nJ)} \\ \midrule
 & LIF & \thead{$(mnFr_{in}+nFr_{out})E_{AC}$\\$ + nE_{MAC}$} &4.52 \\ 
& CE-LIF & \thead{$(mnFr_{in}+nFr_{out})E_{AC}$\\$+3nE_{MAC}$} & 6.43\\ 
FFSNN & LTC & \thead{$(mnFr_{in}+2nFr_{out}+2n)E_{AC}$\\$ + (4nn+3n)E_{MAC}$}&1,210.23 \\ 
& SPSN &\thead{$mnFr_{in}E_{AC}$\\$ + nkE_{MAC} $} &152.14 \\ 
& PMSN &\thead{$mnFr_{in}E_{AC}$\\$ + (2n\log_2(T)+5n)E_{MAC}$} & 30.68\\ \midrule
 & LIF & \thead{$(mnFr_{in}+(nn+n)Fr_{out})E_{AC}$\\$ + nE_{MAC} $}&21.79 \\ 
SRNN & CE-LIF &\thead{$(mnFr_{in}+(nn+n)Fr_{out})E_{AC}$\\$ + 3nE_{MAC} $} &10.96 \\ 
& LTC &\thead{$(mnFr_{in}+(nn+2n)Fr_{out}+2n)E_{AC}$\\$+ (4nn+3n)E_{MAC} $} & 1,211.3\\ \midrule
GSN & & \thead{$(2mnFr_{in}+2nnFr_{out})E_{AC}$\\$ + 5nE_{MAC} $} & 30.48\\ 
Spiking TCN & &\thead{$(kmnFr_{in}+knnFr_{conv2})E_{AC}$} &19.85 \\ 
\makecell[c]{Spike-Driven Transformer} & LIF & \thead{$((3Fr_{in}+Fr_{attn})nn+(Fr_{q}Fr_{k}+Fr_v)nT $\\$ + (Fr_{fc1}+Fr_{fc2})nh )E_{AC}$} & 193.22 \\ 
Binary S4D & & \thead{$2nnFr_{out}E_{AC}$\\$ + (2n\log_2(T)+8n)E_{MAC} $} &65.01 \\ 
GSU & & \thead{$(nn(Fr_{y}+Fr_{w})+2n)E_{AC}$\\$ + (2n\log_2(T)+6n)E_{MAC} $}& 28.36\\ \bottomrule
\end{tabular}}
\label{tab:energy}
\end{table}

\section*{Effectiveness Evaluation on Existing SNN Benchmarks}
In this section, we apply the proposed Segregated Temporal Probe (STP) criterion to evaluate commonly used datasets in the neuromorphic community, shedding light on their effectiveness as benchmarks for assessing temporal processing capacities. For each benchmark, we follow standard configurations for dataset preprocessing, data augmentation, network design, and training setup.

We first examine static image recognition datasets, including MNIST~\cite{lecun1998gradient}, CIFAR10~\cite{krizhevsky2009learning}, and CIFAR100~\cite{krizhevsky2009learning}. These datasets lack temporal dynamics, as each input sequence is generated by repeating a static image across the temporal dimension. This limitation is evidenced by our experimental results. As shown in Table~\ref{tab:res_stp2}, NoTD and SDBP achieve accuracy comparable to STBP, demonstrating that modeling temporal relationships is unnecessary for high performance on these datasets.

Next, we analyze event-based vision datasets, including N-MNIST~\cite{orchard2015converting}, CIFAR10-DVS~\cite{li2017cifar10}, DVS-Gesture~\cite{amir2017low}, DVS-SLR~\cite{zhou2024enhancing}, GAIT-DAY-DVS~\cite{wang2022event}, and HAR-DVS~\cite{wang2024har}. Despite the presence of event-driven dynamics, our results presented in Table~\ref{tab:res_stp3} reveal that NoTD performs similarly to or even outperforms STBP, indicating that these datasets can be effectively addressed at the frame level without the establishment of temporal dependency.

Furthermore, we evaluate keyword spotting datasets, including GSC~\cite{warden2018speech}, SHD~\cite{cramer2020heidelberg}, and SSC~\cite{cramer2020heidelberg}. The results demonstrated in Table~\ref{tab:res_stp4} reveal that effective decisions for these datasets can often be made by integrating short-term temporal features, as indicated by the marginal performance gap between STDP and STBP and the moderate accuracy achieved by NoTD. These findings suggest that temporal propagation primarily contributes to the integration of short-term features during the forward pass, while backward temporal propagation has a negligible impact on establishing temporal dependencies. Consequently, these benchmarks are inadequate for effectively evaluating the temporal processing capacities of SNNs.

These findings underscore the limitations of existing benchmarks and the urgent need for a more effective neuromorphic temporal processing benchmark. NSA addresses these gaps by incorporating tasks that contain rich, multi-timescale temporal dependencies, ensuring an effective and consistent assessment of SNNs' temporal processing capacities.

\begin{table*}[!htb]
\centering
\caption{Validating the effectiveness of static image recognition datasets in evaluating the temporal processing capacity of SNNs.}
\resizebox{0.45\textwidth}{!}{
\begin{tabular}{@{}l|c|c|c@{}}
\toprule
\textbf{Method} & \makecell[c]{\textbf{MNIST}\\Acc. (\%) $\uparrow$ } & \makecell[c]{\textbf{CIFAR10}\\Acc. (\%) $\uparrow$} & \makecell[c]{\textbf{CIFAR100}\\Acc. (\%) $\uparrow$}   \\ \midrule
STBP & \textbf{99.40} & \textbf{94.86}  & \textbf{74.57}  \\
SDBP & 99.27 (-0.13) & 94.74 (-0.12)  & 74.35 (-0.22)    \\
NoTD & 99.18 (-0.22) & 93.46 (-1.40)  & 73.28 (-1.29) \\ \bottomrule
\end{tabular}}
\label{tab:res_stp2}
\end{table*}

\begin{table*}[!htb]
\centering
\caption{Validating the effectiveness of event-based vision datasets in evaluating the temporal processing capacity of SNNs.}
\resizebox{0.9\textwidth}{!}{
\begin{tabular}{@{}l|c|c|c|c|c|c@{}}
\toprule
\textbf{Method} & \makecell[c]{\textbf{N-MNIST}\\Acc. (\%) $\uparrow$} & \makecell[c]{\textbf{CIFAR10-DVS}\\Acc. (\%) $\uparrow$} & \makecell[c]{\textbf{DVS-Gesture}\\\ Acc. (\%) $\uparrow$}& \makecell[c]{\textbf{DVS-SLR}\\Acc. (\%) $\uparrow$} & \makecell[c]{\textbf{GAIT-DAY-DVS}\\Acc. (\%) $\uparrow$} & \makecell[c]{\textbf{HARDVS}\\\ Acc. (\%) $\uparrow$} \\ \midrule
STBP  & \textbf{99.49} & 78.50 &  95.14 & \textbf{76.57} & 93.52 &  \textbf{48.64} \\
SDBP   & 99.48 (-0.01) & 79.00 (+0.50) & \textbf{95.83} (+0.69)  & 75.92 (-0.65) & \textbf{93.70} (+0.18) & 48.01 (-0.63)  \\
NoTD   & 99.09 (-0.40) & \textbf{80.00} (+1.50) & 94.44 (-0.70)  & 75.18 (-1.39) & 93.47 (-0.05) & 47.98 (-0.66)   \\ \bottomrule
\end{tabular}}
\label{tab:res_stp3}
\end{table*}

\begin{table*}[!htb]
\centering
\caption{Validating the effectiveness of keyword spotting datasets in evaluating the temporal processing capacity of SNNs.}
\resizebox{0.50\textwidth}{!}{
\begin{tabular}{@{}l|c|c|c@{}}
\toprule
\textbf{Method} & \makecell[c]{\textbf{GSC}\\Acc. (\%) $\uparrow$}  & \makecell[c]{\textbf{SHD}\\Acc. (\%) $\uparrow$ } & \makecell[c]{\textbf{SSC}\\Acc. (\%) $\uparrow$}    \\ \midrule
STBP & \textbf{92.91} & \textbf{86.48} & \textbf{67.13}   \\
SDBP &89.00 (-3.91) & 85.07 (-1.41) & 66.03 (-1.10)      \\
NoTD  &77.53 (-15.38) & 68.51 (-17.97) & 44.97 (-22.16)    \\ \bottomrule
\end{tabular}}
\label{tab:res_stp4}
\end{table*}

\section*{Experimental Setups}

\subsection*{Computing Infrastructure}
All experiments are conducted on an Ubuntu 20.04.5 LTS server equipped with NVIDIA GeForce RTX 3090 GPUs (24GB), an Intel Xeon Platinum 8370C CPU, PyTorch 1.13.0, and CUDA 11.8.

\subsection*{Data Preprocessing}

\begin{itemize}
    \item \textbf{AL:} In this task, we utilize our proposed synthetic dataset AL. Each sample begins with a robot positioned at coordinate (0, 0), facing the positive y-axis. The robot then executes a randomly generated sequence of actions, with each action sampled from the set \{`turn left', `turn right', `go straight', `stop'\} according to the probabilities [0.05,~0.05,~0.45,~0.45], respectively. At the end of the sequence, the robot's final x-coordinate determines the classification: $x \leq 0$ is ``left'', otherwise ``right''.
    \item \textbf{HAR:} The smartwatch gyroscope data (sampled at 20 Hz) from the WISDM dataset~\cite{weiss2019wisdm} is segmented using a sliding window approach. Each window has a length of 200 time steps, with a hop length of 100 time steps. The data features angular velocity in three coordinates (X, Y, Z). To eliminate subject-specific bias and ensure consistency, data from all subjects is z-score normalized per coordinate, shuffled, and split into training and testing sets.
    \item \textbf{EEG-MI:} EEG motor imagery data from the OpenBMI dataset~\cite{lee2019} is preprocessed using a widely adopted EEG signal preprocessing pipeline \cite{hosseini2020review}. The raw signals are bandpass-filtered (0.5–80 Hz), downsampled to 100 Hz, and cleaned using Independent Component Analysis (ICA) to remove artifacts. The data are then segmented to extract the principal EEG data relevant to the task. Subsequently, each channel is z-score normalized to ensure feature consistency. Finally, the samples are shuffled across subjects and split into training and testing sets to eliminate subject-specific biases and ensure effective model evaluation.
    \item \textbf{SSL:} We preprocess the raw audio signals (sampled at 16 kHz) provided by the SLoClas dataset \cite{qian2021sloclas} using a sliding window approach with a window size of 500 and a hop size of 500. To simulate real-world application scenarios, directional background noise from 0°, 90°, 180°, and 270° provided in the dataset are randomly segmented and added to each sample at 0 SNR. Notably, raw time-domain signals are processed directly, without the application of any frequency spectral analysis.
    \item \textbf{ALR:} We preprocess the raw event data in the DVS-Lip dataset~\cite{tan2022multi} by converting asynchronous events into frame-based representations. Specifically, events are grouped into their nearest temporal bins, forming $T$ temporal bins, where $T=200$ in our experiments. To preserve polarity information, positive and negative events are mapped into two separate channels. For data augmentation, we apply a series of transformations to the training data, including center cropping and random cropping to achieve an $88\times88$ spatial resolution, horizontal flipping with a probability of $0.5$, cutout, and random zoom-in/zoom-out. 
    For the test data, only center cropping to $88\times88$ is applied.
    \item \textbf{AD:} All audio samples in the N-DNS dataset~\cite{timcheck2023intel} are processed at a sampling rate of 16 kHz and standardized to a fixed duration of 30 seconds. For audio clips shorter than 30 seconds, we concatenate additional speech segments from the same speaker, inserting a 0.2-second silence between utterances to maintain natural pauses. Noisy audio is generated by mixing clean speech with randomly selected noise samples at signal-to-noise ratios ranging from -5 dB to 20 dB, ensuring a diverse range of noise conditions. To simulate varying input loudness levels, we apply loudness normalization to each noisy sample, adjusting levels between -35 dB and -15 dB relative to full scale. For spectral analysis, the Short-Time Fourier Transform (STFT) is configured with a 32 ms window length (512 samples) and an 8 ms hop length (128 samples), using a Hanning window and 512-point Fast Fourier Transform to capture detailed frequency information.
    \item \textbf{ASR:} Experiments on the AISHELL dataset~\cite{bu2017aishell} are conducted using the ESPnet~\cite{watanabe2018espnet} toolkit, following its standard recipes for data preparation, model training, and evaluation. This ensures a consistent and reproducible experimental setup. 
\end{itemize}

\subsection*{Task-Specific Configurations}

In all experiments, the output layer of the SNNs is implemented as a non-spiking fully connected readout layer. 
For AL, HAR, and SSL tasks, the final network output is obtained by averaging the readout layer's outputs across all time steps. For EEG-MI and ALR tasks, the network output is derived solely from the readout layer's output at the final time step. For AD and ALR tasks, where labels vary across time steps, the network output at each time step is dynamically determined by the corresponding readout layer's output.

For AL, HAR, EEG-MI, SSL, and ALR tasks, we use the AdamW optimizer with Cross-Entropy (CE) loss. In AD experiments, following~\cite{hao2024towards}, we adopt a combination of magnitude loss and complex loss in the time-frequency domain alongside the Scale-Invariant Signal-to-Distortion Ratio (SI-SDR) loss. For ASR experiments, the Connectionist Temporal Classification (CTC) loss is employed. Additionally, the triangle surrogate function \cite{deng2022temporal} is consistently applied across all experiments.

In \textbf{\nameref{sec:effectivness} (Sec. \ref{sec:effectivness})}, the sequence length of the AL task is adjusted to 200 thereby highlighting the performance gap among the three training algorithms in STP. All experiments in this section follow the hyperparameters and model configurations of the LIF-SFNN models listed in Tables \ref{tab:hyper1} and \ref{tab:hyper2}.

In \textbf{\nameref{sec:performance} (Sec. \ref{sec:performance})}, the default sequence length for the AL task is extended to 400. To ensure fair comparisons, all models are configured with an equivalent number of trainable parameters to the baseline SFNN with LIF neurons for each task. For AL, HAR, EEG-MI, and SSL tasks, the number of trainable parameters is set to approximately 0.1$M$ for all models. For ALR, AD, and ASR tasks, the trainable parameters are approximately 9.5$M$, 0.4$M$, and 4.7$M$, respectively. Detailed hyperparameters and model specifications for all tasks are provided in Tables \ref{tab:hyper1} and \ref{tab:hyper2}.
In terms of model-specific hyperparameters, $\beta$ represents the threshold decay constant for CE-LIF. For SPSN and Spiking TCN models, $k$ denotes the kernel size, which defines the receptive field over time for temporal feature extraction. In the Spike-Driven Transformer, $nhead$ specifies the number of attention heads in the multi-head attention mechanism. Additionally, $wd$ is an abbreviation for weight decay, which is utilized to regularize spatial credit assignment in Spiking S4D and GSU models.

In \textbf{\nameref{sec:efficiency} (Sec. \ref{sec:efficiency})}, we evaluate training speed and memory cost using sequence lengths of 200, 400, and 800 on the AL task, while setting the sequence length to 400 for energy efficiency evaluation. All SNN models are standardized with two hidden layers of 256 channels each, ensuring consistent hidden dimensions for fair comparisons. The batch size is fixed at 256, and other model-specific hyperparameters remain unchanged from prior experiments.

\begin{table*}[htbp]
\centering
\caption{Hyperparameter configurations and model specifications for AL, HAR, EEG-MI, and SSL tasks.}
\label{tab:hyper1}
\resizebox{0.98\textwidth}{!}{%
\begin{tabular}{@{}l|l|
l|cccccccc|c@{}}
\toprule
\textbf{Task} & \textbf{Architecture} & \textbf{Neuron model} & \makecell[c]{\textbf{Batch}\\\textbf{size}} & \textbf{Epochs} & \makecell[c]{\textbf{Hidden}\\\textbf{dimension}} & \makecell[c]{\textbf{Learning}\\\textbf{rate}} & \makecell[c]{\textbf{Grad}\\\textbf{clipping}} & \textbf{Decay} & \textbf{Threshold} & \makecell[c]{\textbf{Surrogate}\\ \textbf{gradient width}} & \textbf{Others} \\ \midrule
 
 \multirow{15}{*}{\bf AL} 
 & \multirow{5}{*}{FFSNN} & LIF & 256 & 100 & 128-256-256 & 5.00E-03 & 1 & 1 & 0.5 & 0.4 & \\ 
 &  & CE-LIF & 256 & 100 & 128-152-152 & 3.00E-03 & 1 & 0.5 & 0.5 & 0.4 & $\beta$=0.02 \\ 
 &  & LTC & 256 & 100 & 96-128 & 1.00E-03 & 1 & - & 0.5 & 0.4 & \\ 
 &  & SPSN & 256 & 100 & 128-256-256 & 5.00E-03 & 1 & - & 0.5 & 0.4 & $k$=128 \\ 
 &  & PMSN & 256 & 100 & 64-256-256 & 2.00E-02 & 1 & - & 0.5 & 1 & \\ 
 \cmidrule{2-12}
 & \multirow{3}{*}{SRNN} & LIF & 256 & 100 & 128-176-176 & 1.00E-03 & 1 & 0.5 & 0.5 & 0.4 & \\ 
 &  & CE-LIF & 256 & 100 & 120-120-120 & 1.00E-03 & 1 & 0.3 & 0.5 & 0.4 & $\beta$=0.03 \\ 
 &  & LTC & 256 & 100 & 96-112 & 2.00E-03 & 0.25 & - & 0.5 & 0.4 & \\ 
  \cmidrule{2-12}
 & GSN &  & 256 & 100 & 64-112-112 & 3.00E-03 & 1 & - & 0.5 & 0.6 & \\ 
 & Spiking TCN &  & 256 & 100 & 36*6 & 5.00E-03 & 1 & 0.5 & 0.5 & 1 & $k$=7 \\ 
 & Spike-Driven Transformer & LIF & 256 & 100 & 64-64 & 5.00E-04 & 1 & 0.5 & 0.5 & 0.6 & $nhead$=8 \\ 
 & Binary S4D &  & 256 & 100 & 128-128 & 3.00E-03 & 1 & - & 0.5 & 0.6 & $wd$=3.00E-03 \\ 
 & GSU &  & 256 & 100 & 152-152 & 5.00E-03 & 1 & - & 0.5 & 0.6 & $wd$=3.00E-03 \\ 

  \midrule

\multirow{15}{*}{\bf HAR} 
 & \multirow{5}{*}{FFSNN} & LIF & 256 & 100 & 128-256-256 & 3.00E-03 & 1 & 0.9 & 0.5 & 0.6 & \\ 
 &  & CE-LIF & 256 & 100 & 128-192-192 & 3.00E-03 & 1 & 0.3 & 0.5 & 0.6 & $\beta$=0.15 \\ 
 &  & LTC & 256 & 100 & 96-128 & 3.00E-03 & 1 & - & 0.5 & 0.6 & \\ 
 &  & SPSN & 256 & 100 & 128-256-256 & 3.00E-03 & 1 & - & 0.5 & 0.6 & $k$=16 \\ 
 &  & PMSN & 256 & 100 & 64-256-256 & 5.00E-03 & 1 & - & 0.5 & 1 & \\ 
 \cmidrule{2-12}
 & \multirow{3}{*}{SRNN} & LIF & 256 & 100 & 128-176-176 & 1.50E-03 & 0 & 0.1 & 0.5 & 0.6 & \\ 
 &  & CE-LIF & 256 & 100 & 128-128-160 & 3.00E-03 & 1 & 0.3 & 0.5 & 0.6 & $\beta$=0.1 \\ 
 &  & LTC & 256 & 100 & 96-112 & 3.00E-03 & 1 & - & 0.5 & 0.6 & \\ 
  \cmidrule{2-12}
 & GSN &  & 256 & 100 & 64-112-112 & 5.00E-03 & 1 & - & 0.5 & 0.6 & \\ 
 & Spiking TCN &  & 256 & 100 & 76 & 5.00E-03 & 1 & 0.5 & 0.5 & 1 & $k$=16 \\ 
 & Spike-Driven Transformer & LIF & 256 & 100 & 64-64 & 3.00E-03 & 1 & 0.5 & 0.6 & 0.8 & $nhead$=1 \\ 
 & Binary S4D &  & 256 & 100 & 128-128 & 1.00E-02 & 1 & - & 0.5 & 0.6 & $wd$=3.00E-03 \\ 
 & GSU &  & 256 & 100 & 152-152 & 1.00E-02 & 1 & - & 0.5 & 0.6 & $wd$=3.00E-03 \\ 
 
\midrule
 
\multirow{15}{*}{\bf EEG-MI} 
 & \multirow{5}{*}{FFSNN} & LIF & 128 & 100 & 128-256-256 & 1.00E-03 & 1 & 1 & 0.5 & 0.6 & \\ 
 &  & CE-LIF & 128 & 100 & 128-128-128 & 1.00E-03 & 1 & 0.3 & 0.5 & 0.6 & $\beta$=0.02 \\ 
 &  & LTC & 128 & 100 & 96-112 & 1.00E-03 & 0 & - & 0.5 & 0.6 & \\ 
 &  & SPSN & 128 & 100 & 128-256-256 & 3.00E-03 & 1 & - & 0.5 & 0.6 & $k$=500 \\ 
 &  & PMSN & 128 & 100 & 64-256-256 & 1.00E-02 & 1 & - & 0.5 & 1 & \\ 
 \cmidrule{2-12}
 & \multirow{3}{*}{SRNN} & LIF & 128 & 100 & 128-176-176 & 2.00E-04 & 1 & 1 & 0.5 & 0.6 & \\ 
 &  & CE-LIF & 128 & 100 & 64-128-128 & 2.00E-04 & 1 & 0.2 & 0.5 & 0.6 & $\beta$=0.02 \\ 
 &  & LTC & 128 & 100 & 96-104 & 1.00E-03 & 0 & - & 0.5 & 0.6 & \\ 
  \cmidrule{2-12}
 & GSN &  & 128 & 100 & 64-112-112 & 5.00E-03 & 1 & - & 0.5 & 0.6 & \\ 
 & Spiking TCN &  & 128 & 100 & 28*8 & 3.00E-03 & 1 & 0.5 & 0.5 & 1 & $k$=7 \\ 
 & Spike-Driven Transformer & LIF & 128 & 100 & 64-64 & 5.00E-04 & 1 & 0.5 & 0.5 & 0.6 & $nhead$=8 \\ 
 & Binary S4D &  & 128 & 100 & 128-128 & 1.00E-02 & 1 & - & 0.5 & 0.6 & $wd$=5.00E-03 \\ 
 & GSU &  & 128 & 100 & 152-152 & 1.00E-02 & 1 & - & 0.5 & 0.6 & $wd$=5.00E-03 \\ 
 
 \midrule
 
 \multirow{15}{*}{\bf SSL} 
 & \multirow{5}{*}{FFSNN} & LIF & 64 & 100 & 128-256-256 & 5.00E-03 & 1 & 0.6 & 0.5 & 0.6 & \\ 
 &  & CE-LIF & 64 & 100 & 64-128-128 & 5.00E-03 & 1 & 0.3 & 0.5 & 0.6 & $\beta$=0.05 \\ 
 &  & LTC & 64 & 100 & 96-112 & 3.00E-03 & 1 & - & 0.5 & 0.6 & \\ 
 &  & SPSN & 64 & 100 & 128-256-256 & 3.00E-03 & 1 & - & 0.5 & 0.6 & $k$=16 \\ 
 &  & PMSN & 64 & 100 & 64-256-256 & 3.00E-03 & 1 & - & 0.5 & 1 & \\ 
 \cmidrule{2-12}
 & \multirow{3}{*}{SRNN} & LIF & 64 & 100 & 128-168-168 & 5.00E-04 & 1 & 0.5 & 0.5 & 0.4 & \\ 
 &  & CE-LIF & 64 & 100 & 96-108-108 & 1.00E-03 & 1 & 0.3 & 0.5 & 0.6 & $\beta$=0.02 \\ 
 &  & LTC & 64 & 100 & 96-104 & 1.00E-03 & 0.5 & - & 0.5 & 0.6 & \\ 
  \cmidrule{2-12}
 & GSN &  & 64 & 100 & 96-112-112 & 1.00E-03 & 0.15 & - & 0.5 & 0.4 & \\ 
 & Spiking TCN &  & 64 & 100 & 46*8 & 3.00E-03 & 1 & 0.5 & 0.5 & 1 & $k$=3 \\ 
 & Spike-Driven Transformer & LIF & 64 & 100 & 64-64 & 5.00E-04 & 1 & 0.5 & 0.5 & 0.6 & $nhead$=4 \\ 
 & Binary S4D &  & 64 & 100 & 128-128 & 1.00E-02 & 1 & - & 0.5 & 0.6 & $wd$=5.00E-03 \\ 
 & GSU &  & 64 & 100 & 152-152 & 1.00E-02 & 1 & - & 0.5 & 0.6 & $wd$=5.00E-03 \\

 \bottomrule
\end{tabular}}
\end{table*}

\begin{table*}[htbp]
\centering
\caption{Hyperparameter configurations and model specifications for ALR, AD, and ASR tasks.}
\label{tab:hyper2}
\resizebox{0.98\textwidth}{!}{%
\begin{tabular}{@{}l|l|
l|cccccccc|c@{}}
\toprule
\textbf{Task} & \textbf{Architecture} & \textbf{Neuron model} & \makecell[c]{\textbf{Batch}\\\textbf{size}} & \textbf{Epochs} & \makecell[c]{\textbf{Hidden}\\\textbf{dimension}} & \makecell[c]{\textbf{Learning}\\\textbf{rate}} & \makecell[c]{\textbf{Grad}\\\textbf{clipping}} & \textbf{Decay} & \textbf{Threshold} & \makecell[c]{\textbf{Surrogate}\\ \textbf{gradient width}} & \textbf{Others} \\ \midrule
\multirow{15}{*}{\bf ALR} 
 & \multirow{5}{*}{FFSNN} & LIF  & 256 & 100 & 512*6 & 3.00E-03 & 0 & 0.95 & 0.8 & 1 & \\ 
                      &  & CE-LIF & 256 & 100 & 512*6 & 3.00E-03 & 1 & 0.3 & 0.5 & 1 & $\beta$=0.02 \\ 
                        &  & LTC & 256 & 100 & 512*6 & 3.00E-03 & 0 & - & 0.8 & 1 & \\ 
                       &  & SPSN & 256 & 100 & 512*6 & 3.00E-03 & 0 & - & 0.8 & 1 & $k$=32 \\ 
                       &  & PMSN & 256 & 100 & 512*6 & 3.00E-03 & 0 & - & 0.8 & 1 & \\ 
 \cmidrule{2-12}
   & \multirow{3}{*}{SRNN} & LIF & 256 & 100 & 460*6 & 5.00E-04 & 0 & 0.95 & 0.8 & 1 & \\ 
                      &  & CE-LIF & 256 & 100 & 460*6 & 3.00E-03 & 1 & 0.3 & 0.5 & 1 & $\beta$=0.02 \\ 
                        &  & LTC & 256 & 100 & 340*6 & 5.00E-04 & 0 & - & 0.8 & 1 & \\ 
  \cmidrule{2-12}
                    & GSN &      & 256 & 100 & 256*6 & 5.00E-04 & 0 & - & 0.1 & 1 & \\ 
                & Spiking TCN &  & 256 & 100 & 75*6 & 3.00E-03  & 0 & 0.5 & 0.5 & 1 & $k$=7 \\ 
& Spike-Driven Transformer & LIF & 256 & 100 & 272*6 & 5.00E-04 & 0 & 0.5 & 0.8 & 1   & $nhead$=4 \\ 
                 & Binary S4D &  & 256 & 100 & 435*6 & 3.00E-03 & 0 & - & 0.8 & 1 & $wd$=5.00E-04 \\ 
                    & GSU &  & 256 & 100 & 485*6 & 3.00E-03 & 0 & - & 0.8 & 1 & $wd$=5.00E-04\\ 
 
\midrule
 
\multirow{15}{*}{\bf AD} 
 & \multirow{4}{*}{FFSNN} & LIF  & 64 & 50 & 320*2/224*2 & 1.00E-03 & 10 & 0.8 & 0.5 & 1 & \\ 
                      
                        &  & LTC & 64 & 50 & 140*2/85*2  & 1.00E-03 & 10 & - & 0.5 & 1 & \\ 
                       &  & SPSN & 64 & 50 & 320*2/224*2 & 1.00E-03 & 10 & - & 0.5 & 1 & $k$=32 \\ 
                       &  & PMSN & 64 & 50 & 320*2/224*2 & 5.00E-04 & 10 & - & 0.5 & 1 & \\ 
 \cmidrule{2-12}
   & \multirow{2}{*}{SRNN} & LIF & 64 & 50 & 280*2/160*2 & 1.00E-03 & 0 & 0.5 & 0.5 & 1 & \\  
                        &  & LTC & 64 & 50 & 136*2/80*2  & 1.00E-03 & 10 & - & 1.0 & 1 & \\ 
  \cmidrule{2-12}
                    & GSN &      & 64 & 50 & 155*2/100*2 & 1.00E-03 & 10 & - & 0.5 & 1 & \\ 
                & Spiking TCN &  & 64 & 50 & 125*2/75*2  & 1.00E-03 & 10 & 0.5 & 0.5 & 1 & $k$=3  \\ 
& Spike-Driven Transformer & LIF & 64 & 50 & 84*2/56*2   & 5.00E-04 & 10 & 0.5 & 0.5 & 1   & $nhead$=4 \\ 
                 & Binary S4D &  & 64 & 50 & 170*2/100*2 & 1.00E-03 & 10 & - & 0.5 & 1 &  \\ 
                    & GSU &  & 64 & 50 & 195*2/130*2 & 1.00E-03 & 10 & - & 0.5 & 1 & \\ 

  \midrule
 
\multirow{15}{*}{\bf ASR} 
 & \multirow{5}{*}{FFSNN} & LIF  & 360 & 20 & 1024*3 & 2 & 0 & 0.95 & 0.8 & 1 & \\ 
                      &  & CE-LIF & 360 & 20 & 1024*3 & 2 & 0 & 0.5 & 0.5 & 0.6 & $\beta$=0.2 \\ 
                        &  & LTC & 360 & 20 & 512*3 & 2 & 0 & - & 0.8 & 1 & \\ 
                       &  & SPSN & 360 & 20 & 1024*3 & 2 & 0 & - & 0.5 & 1 & $k$=32 \\ 
                       &  & PMSN & 360 & 20 & 1024*3 & 1 & 0 & - & 0.5 & 1 & \\ 
 \cmidrule{2-12}
   & \multirow{3}{*}{SRNN} & LIF & 360 & 20 & 820*3 & 2 & 0 & 0.3 & 0.5 & 1 & \\ 
                      &  & CE-LIF & 360 & 20 & 770*3 & 2 & 0 & 0.3 & 0.5 & 0.6 & $\beta$=0.2 \\ 
                        &  & LTC & 360 & 20 & 480*3 & 2 & 0 & - & 1.0 & 0.8 & \\ 
  \cmidrule{2-12}
                    & GSN &      & 360 & 20 & 512*3 & 2 & 0 & - & 0.8 & 0.6 & \\ 
                & Spiking TCN &  & 360 & 20 & 128*3 & 2& 0 & 0.5 & 0.5 & 1 & $k$=7 \\ 
& Spike-Driven Transformer & LIF & 360 & 20 & 328*3 & 1 & 0 & 0.5 & 0.3 & 1   & $nhead$=4 \\ 
                 & Binary S4D &  & 360 & 20 & 680*3 & 2 & 0 & - & 0.5 & 0.6 &  \\ 
                    & GSU &  & 360 & 20 & 850*3 & 2 & 0 & - & 0.5 & 0.6 & \\

 \bottomrule
\end{tabular}}
\end{table*}

\end{document}